\RequirePackage{fix-cm}
\UseRawInputEncoding 
\documentclass[twocolumn]{svjour3}          
\smartqed  
\usepackage{graphicx}
\usepackage{afterpage}
\usepackage[figuresright]{rotating}
\usepackage{adjustbox}
\usepackage[round]{natbib}
\usepackage[doipre={DOI:}]{uri}
\usepackage{subfigure}
\usepackage{multirow}
\usepackage{booktabs}
\usepackage{xcolor}
\usepackage{amsmath,amssymb,amsfonts}
\usepackage{verbatim}
\usepackage[pagebackref=true,breaklinks=true,colorlinks,bookmarks=false,citecolor=blue,linkcolor=blue]{hyperref}
\usepackage{url}
\usepackage{tabularx}
\usepackage{lscape}

\usepackage{enumitem}
\usepackage{bm}
\usepackage{amssymb}
\usepackage{caption}
\usepackage{graphicx}
\usepackage{subfigure}
\usepackage{bbm}
\usepackage{colortbl}
\usepackage{mathtools}
\usepackage{makecell}
\newcommand{\RNum}[1]{\uppercase\expandafter{\romannumeral #1\relax}}
\usepackage{bbding}

\newcolumntype{C}[1]{>{\centering\let\newline\\\arraybackslash\hspace{0pt}}m{#1}} 
\newcolumntype{L}[1]{>{\let\newline\\\arraybackslash\hspace{0pt}}m{#1}} 

\usepackage{booktabs,lscape}

\begin{document}
\begin{sloppypar}

\title{DocScanner: Robust Document Image Rectification with Progressive Learning
}

\author{Hao Feng  \and
        Wengang Zhou \and
        Jiajun Deng \and
        Qi Tian \and
        Houqiang Li
}

\institute{
            \\
            Hao Feng \at
            University of Science and Technology of China\\
              \email{fh1995@mail.ustc.edu.cn}
              \and\\
            Wengang Zhou \at
            University of Science and Technology of China\\
              \email{zhwg@ustc.edu.cn}
              \and\\
            Jiajun Deng \at
            The University of Sydney\\
              \email{jiajun.deng@sydney.edu.au}
              \and\\
            Qi Tian \at
            Huawei Cloud $\&$ AI\\
              \email{tian.qi1@huawei.com}
              \and\\
            Houqiang Li \at
            University of Science and Technology of China\\
             \email{lihq@ustc.edu.cn}
}

\date{Received: date / Accepted: date}

\maketitle

\begin{abstract}
Compared with flatbed scanners, portable smartphones provide more convenience for physical document digitization. 
However, such digitized documents are often distorted due to uncontrolled physical deformations, camera positions, and illumination variations. 
To this end, we present DocScanner, a novel framework for document image rectification. Different from existing solutions, DocScanner addresses this issue by introducing a progressive learning mechanism. 
Specifically, DocScanner maintains a single estimate of the rectified image, which is progressively corrected with a recurrent architecture. The iterative refinements make DocScanner converge to a robust and superior rectification performance, while the lightweight recurrent architecture ensures the running efficiency.
To further improve the rectification quality, based on the geometric priori between the distorted and the rectified images, a geometric regularization is introduced during training to further improve the performance.
Extensive experiments are conducted on the Doc3D dataset and the DocUNet Benchmark dataset, and the quantitative and qualitative evaluation results verify the effectiveness of DocScanner, which outperforms previous methods on OCR accuracy, image similarity, and our proposed distortion metric by a considerable margin. 
Furthermore, our DocScanner shows superior efficiency in runtime latency and model size.
\end{abstract}

\section{Introduction}
Document digitization refers to the creation of a digital image backup of a document file, which is frequently applied in many formal affairs. Thanks to the rapid advances in portable cameras and smartphones, document digitization becomes much more accessible than before. However, such captured document images commonly suffer from various levels of distortions, due to uncontrolled camera position, uneven illumination, and various paper sheet deformations (\emph{i.e.}, folded, curved, and crumpled). These distortions make the digital files unqualified on many occasions. Besides, they also bring difficulties to many downstream processings, such as automatic text recognition~\cite{yuan2022syntax,peng2022pagenet}, content understanding~\cite{zhong2019publaynet,kim2022ocr}, and question answering~\cite{mathew2021docvqa}, editing, and preservation. To address these problems, document image rectification has been actively researched in recent years.

One direction of the early attempts to document image rectification is developed based on the reconstruction of the 3D shape of deformed pages. Those methods heavily rely on auxiliary hardware \cite{937649, brown2007,4407722, 6909892} or multiview shooting \cite{brown2004image, yamashita2004shape, 4916075, 7866848}, limiting their further applications. Some other methods~\cite{958227,wu2002document,1561180,6628653} assume a parametric model on the curved pages and optimize the model with specific attributes, such as shading, boundaries, and textlines. Nevertheless, the oversimplified parametric models of such approaches usually lead to limited performance as well as non-negligible computational costs for model optimization.

Recently, deep learning has been introduced to document image rectification with promising performance as well as a significant reduction in computational cost. In such methods~\cite{8578592,9010747,li2019document,liu2020geometric,xie2020dewarping,das2020intrinsic,markovitz2020can,das2021end, feng2021doctr,DDCPXIE,feng2022docgeonet,zhang2022marior,jiang2022revisiting,ma2022learning}, document image rectification is approached 
as the regression of
a dense 2D vector field (warping flow) that samples the pixels from distorted images to rectified ones. 
Typically,
DocUNet~\cite{8578592} first demonstrates the potential of deep learning for document image rectification with a stacked U-Net~\cite{ronneberger2015u}.
Then, DewarpNet~\cite{9010747} models the 3D shape of a deformed document in the network, while DocGeoNet~\cite{feng2022docgeonet} and RDGR~\cite{jiang2022revisiting} leverage the curved textlines to guide the rectification.
DocProj~\cite{li2019document} and PWUNet~\cite{das2021end} consider distinct local deformation fields and stitch them together to obtain an improved restoration.
Although they report superior performance on the challenging benchmark dataset~\cite{8578592}, 
the rectified images remain unsolved distortions and these advanced solutions are still limited by difficulties such as large sheet deformations and distorted textlines. 
In contrast, we propose to conduct distortion rectification in a progressive manner, 
aiming to obtain a robust and superior rectification result.

In this work, we introduce DocScanner, a novel deep network framework for document image rectification.
Different from existing solutions, DocScanner approaches the task by introducing a progressive learning mechanism.
Specifically, 
DocScanner takes a recurrent structure that corrects the document distortion via iterative and progressive refinements.
During training, at each iteration, DocScanner takes the rectification results of the previous iteration as input, aggregates them, and learns to refine the current rectified image toward a distortion-free one.
Note that such a refinement operation can be applied iteratively during inference without divergence.
In this way, the distortions in the input document images are progressively corrected and finally converge to a relatively steady status, achieving an accurate and robust rectification.

Our DocScanner exhibits a novel design, discussed next. 
Firstly, our recurrent rectification architecture maintains a single estimate of the rectified image that is refined iteratively. This is different from the intuitive strategy that a rectified image pyramid is supervised to refine the output in a multi-scale way~\cite{zamir2021multi,yang2021progressively}, where large degradations/deformations are recovered at low resolution, while small ones are recovered at high resolution, which may have difficulty in recovering from early errors.
Secondly, at each iteration, DocScanner aggregates the results predicted at the previous iteration, including the features of the original distorted image and the current rectified image. Then, a convolution-based gated recurrent unit takes the aggregated features and the current hidden state as input, and outputs the refined rectified image.
Thirdly, the recurrent architecture is lightweight with only 4.1M parameters, which ensures efficiency under multiple iterations.
Fourthly, we propose a circle-consistency loss as a geometric regularization to further relieve the rectified distortion, which imposes straight-line constraints on rectified images.

Moreover, we propose a new evaluation metric for document image rectification.
Based on the dense SIFT-flow~\cite{5551153} between ground truth image to rectified one,
the typical metric Local Distortion (LD)~\cite{7866848} computes the average displacement of all matched pixels.
We observe that LD focuses more on the distortion of local areas.
Inspired by this,
we propose Line Distortion (Li-D) as a supplementary metric to further evaluate the global distortion of the rectified images,
by computing the average deformation of the row and column pixels in rectified images.

Extensive experiments on the Doc3D dataset~\cite{9010747} and DocUNet Benchmark dataset~\cite{8578592} demonstrate the effectiveness of our method as well as its superiority over state-of-the-art methods. In addition, we validate various design choices of DocScanner through extensive ablation studies. We summarize the strengths of DocScanner as follows,

\begin{itemize}
   \item \emph{State-of-the-art performance}: DocScanner sets several state-of-the-art records on the challenging DocUNet Benchmark dataset~\cite{8578592}, including metrics MultiScale Structural SIMilarity (MS-SSIM), Local Distortion (LD), our proposed Line Distortion (Li-D), Edit Distance (ED), and Character Error Rate (CER). 

   \item \emph{Superior efficiency}: DocScanner processes 1080P document images at 10.03 FPS on a 2080Ti GPU. Moreover, the parameter number of DocScanner is about \textbf{1/5} of the best-published method.
	
   \item \emph{Strong generalization ability}: DocScanner exhibits strong generalization ability, demonstrated by the robustness experiments on background, viewpoint, illumination, and document type. 
\end{itemize}

\begin{figure*}[t]
	\begin{center}
		\includegraphics[width=1\linewidth]{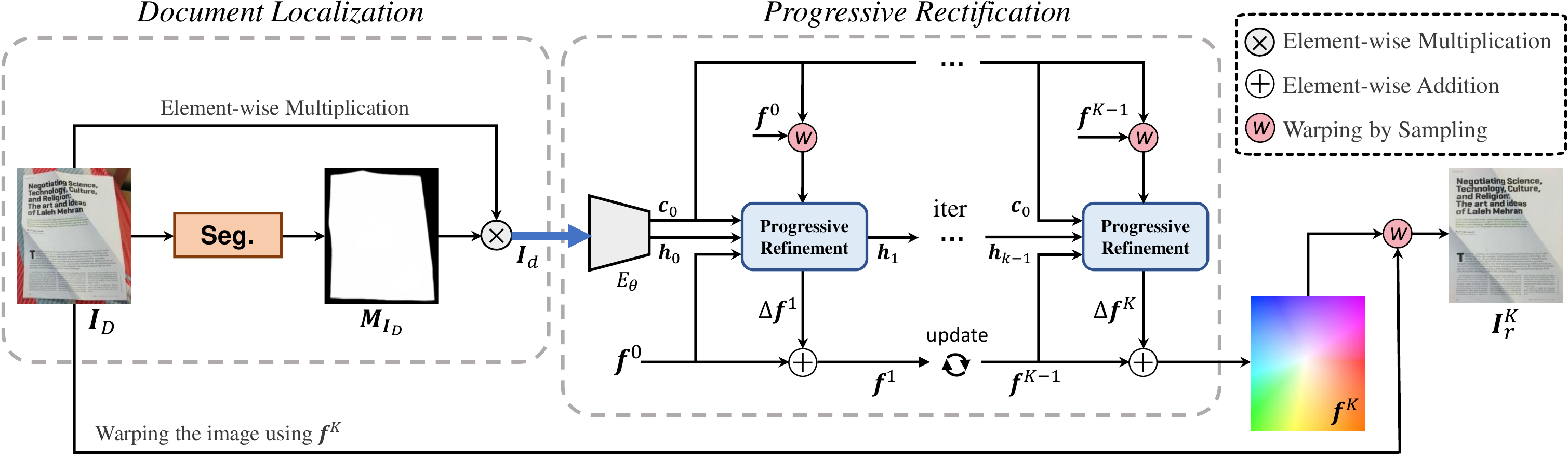}
	\end{center}
	\vspace{-0.12in}
	\caption{\textbf{An overview of the proposed DocScanner.} It decouples the task into a foreground document localization and a geometric distortion rectification. Given the distorted document image $\bm{I}_D$, the document localization module first separates the foreground document from the noisy background by predicting a binary mask $\bm{M}_{\bm{I}_D}$ of the foreground document. Then, the background-excluded image $\bm{I}_d$ is fed into the progressive rectification module, which progressively corrects the geometric distortion in an iterative manner. It maintains a single estimate of the warping flow that is refined at each iteration, and output the rectified image by warping the image $\bm{I}_D$ with the output warping flow $\bm{f}^K$ at the last iteration.}
	\label{fig:2}
	\vspace{-0.11in}
\end{figure*}

\vspace{-0.1in}
\setlength{\parskip}{0em}
\section{Related Work}
In this section, we broadly categorize the research on document image rectification into two different directions: (a) rectification by 3D shape reconstruction, and (b) rectification based on deep learning. In the following, we discuss them separately.

\smallskip
\noindent
\textbf{Rectification by 3D shape reconstruction.} 
Traditionally, some methods utilize auxiliary equipments to reconstruct a 3D shape of the deformed document, followed by flattening the surface to correct the distortions. 
Brown and Seales~\cite{937649} acquire the 3D representation of document shape with a light projector and then flatten this representation via a mass-spring particle system.
Later, they~\cite{brown2007} acquire a 3D scan of the document surface with a 3D scanning system, and conformal mapping~\cite{levy2002least} is used to rectify the geometric distortion by mapping the 3D surface to a plane. 
Zhang \emph{et al}~\cite{4407722} use a laser range scanner and perform restoration by using a physical modeling technique. 
Meng \emph{et al}~\cite{6909892,meng2017active} introduce structured beams illuminating upon the deformed document page to recover curves of the page surface.

In addition to the methods that rely on the auxiliary hardwares, some other methods utilize two or more multiview images for 3D shape reconstruction.
Brown \emph{et al}~\cite{brown2004image} utilize a calibrated mirror system to obtain the 3D surface using multiview stereo.
Yamashita \emph{et al}~\cite{yamashita2004shape} detect the stereo corresponding points between two images based on the normalization cross correlation method.
Tsoi \emph{et al}~\cite{tsoi2007multi} transform the multiple views into a common coordinate frame based on the document boundaries to correct the distortions. 
Koo \emph{et al}~\cite{4916075} estimate the unfolded surface by the corresponding points between two images by SIFT~\cite{lowe2004distinctive}.
You \emph{et al}~\cite{7866848} propose a ridge-aware 3D reconstruction method to rectify a paper sheet from a few of images.
However, both the auxiliary equipments and multiple images are unavailable in common situations, which limits their applicability.   

Moreover, techniques of the third subcategory reconstruct the 3D shape from a single view.
They commonly assume a parametric model on the document surface, like a cylindrical surface, and fit the model based on the extraction of specific representations.
Among them,
some methods~\cite{wada1997shape, 1561180, courteille2007shape, zhang2009unified} obtain a document shape based on shape from shading technique. 
Some algorithms are designed based on the priori that the textlines are horizontally or vertically aligned in well-rectified images, thus the distorted images can be corrected based on the detection of textlines. 
In early work, the detected textlines are modeled as cubic B-splines by Lavialle \emph{et al}~\cite{958227, meng2011metric}, non-linear curve by Wu and Agam~\cite{wu2002document}, and polynomial approximation by~\cite{mischke2005document, kim2015document, kil2017robust}.
Moreover, features about boundaries~\cite{brown2006geometric,6628653}, characters~\cite{zandifar2007unwarping}, interline spacing and textline orientation~\cite{koo2010state} are extracted to estimate the rectification.
Liang \emph{et al}~\cite{liang2008geometric} estimate the 3D document shape from texture flow information obtained directly from the image.
Tian \emph{et al}~\cite{tian2011rectification} compute the 3D deformation up to a scale factor using SVD.
Meng \emph{et al}~\cite{meng2018exploiting} estimate the 3D shape model through weighted majority voting on the vector fields.
Das \emph{et al}~\cite{9010747} innovatively model the 3D shape of a document with a convolutional network and then regress the warping flow for rectification.

\smallskip
\noindent
\textbf{Rectification based on deep learning.} 
Although the above methods achieve encouraging results, the strong assumptions on surface geometry, contents, and illumination limit their applicabilities.
DocUNet~\cite{8578592} is the first model to demonstrate the potential of deep learning for document image rectification. It predicts a dense forward warping flow with a stacked U-Net~\cite{ronneberger2015u} to unwarp the distorted document image.
DocProj~\cite{li2019document} predicts the warping flow of the cropped distorted document image patches first, rather than the entire image, and then stitches them together to generate a fully rectified image.  
However, the estimation and subsequent stitching of the warping flow patches heavily increase the computational cost.
AGUN~\cite{liu2020geometric} develops a pyramid encoder-decoder architecture, which predicts the forward warping flow at multiple resolutions in a coarse-to-fine fashion.
However, directly feeding the distorted images with complex backgrounds to the network for rectification estimation is difficult, due to the involvement of extra implicit learning to identify the foreground document.
Based on Fully Convolutional Network~\cite{long2015fully}, Xie \emph{et al}~\cite{xie2020dewarping} perform a foreground/background classification as a post-processing to refine the predicted forward warping flow on boundary regions of the document.
To learn a powerful representation for the document image, DocTr~\cite{feng2021doctr} first introduces the self-attention mechanism~\cite{Vaswani2017AttentionIA} from the natural language processing tasks to the filed.
To improve the running efficiency, DDCP~\cite{DDCPXIE} only estimates several pairs of control points to conduct rectification.
PWUNet~\cite{das2021end} concentrates on the distinct distortion of local regions for improved global rectification.
DocGeoNet~\cite{feng2022docgeonet} extracts global and local geometric representations to improve rectification, by the prediction of 3D shape and textlines.
To extract the structural information of a deformed document, FDRNet~\cite{xue2022fourier} focuses on high-frequency components in the Fourier space to improve restoration.
Marior~\cite{zhang2022marior} considers the rectification of the document images with large background regions and gradually rectifies them to a robust state.
RDGR~\cite{jiang2022revisiting} first detect textlines and boundaries in a document image, and then perform the rectification by solving an optimization problem with the proposed grid regularization.
To improve the generalization ability of the network, PaperEdge~\cite{ma2022learning} involves the real-world document images in the training.

Although the field of document image rectification has witnessed rapid progress in recent years, the rectified results of such advanced methods still remain unsolved distortions and are unsatisfactory.
In this work, we propose DocScanner, a new deep architecture for document image rectification, aiming to achieve an accurate and robust distortion rectification.

\section{METHODOLOGY}
In this section, we present our design of DocScanner to facilitate the geometric correction of distorted document images. As shown in Fig. \ref{fig:2}, DocScanner consists of a document localization module and a progressive rectification module. Given a distorted document image $\bm{I}_D$, the document localization module estimates a foreground mask $\bm{M}_{\bm{I}_D}$ to exclude the background. Then, the image with only foreground document $\bm{I}_d$ is fed into the progressive rectification module, which maintains a single estimate of warping flow and refines it across $K$ iterations. The final output warping flow $\bm{f}^K$ is used to rectify the input image $\bm{I}_D$. 
Additionally, to further relieve the distortion of rectified images, we propose a regularization loss to regularize the training of the progressive rectification module. In the following, we elaborate the key components of DocScanner, including the document localization module, the progressive rectification module, and the training strategy.

\vspace{-0.07in}
\subsection{Document Localization Module}\label{DLM}
The goal of the document localization module is to remove the noisy background. It makes the subsequent rectification network focuses on geometric rectification, without extra learning on localizing the document. Following prior work~\cite{DDCPXIE,feng2021doctr,zhang2022marior}, we formulate the foreground document segmentation as a saliency detection problem, and address it with a nested U-structure network~\cite{Qin_2020_PR}. As shown in Fig.~\ref{fig:2}, given a distorted document image $\bm{I}_D \in \mathbb{R}^{H \times W \times 3}$, where $H$ and $W$ are the height and width of the image, we predict a confidence map of the foreground document. This map is further binarized with a threshold $\tau$ to obtain the binary document region mask $\bm{M}_{\bm{I}_D}$. Then, the background of $\bm{I}_D$ can be removed by element-wise matrix multiplication with broadcasting along the channels of $\bm{I}_D$. It should be noted that this module can also be replaced with other alternative segmentation networks. 
The document localization module is trained with a binary cross-entropy loss~\cite{de2005tutorial} as follows,
\begin{equation}
	\mathcal{L}_{seg} = -\sum_{i=1}^{N_p}\left[y_i\log(\hat{p_i})+(1-y_i)\log(1-\hat{p_i})\right],
\end{equation}
where $N_p$ is the number of the pixels of the distorted image $\bm{I}_D$, $y_i \in \{0, 1\}$ and $\hat{p_i} \in [0, 1]$ denote the ground-truth and the predicted confidence, respectively.

\begin{figure}[t]
	\begin{center}
		\includegraphics[width=1\linewidth]{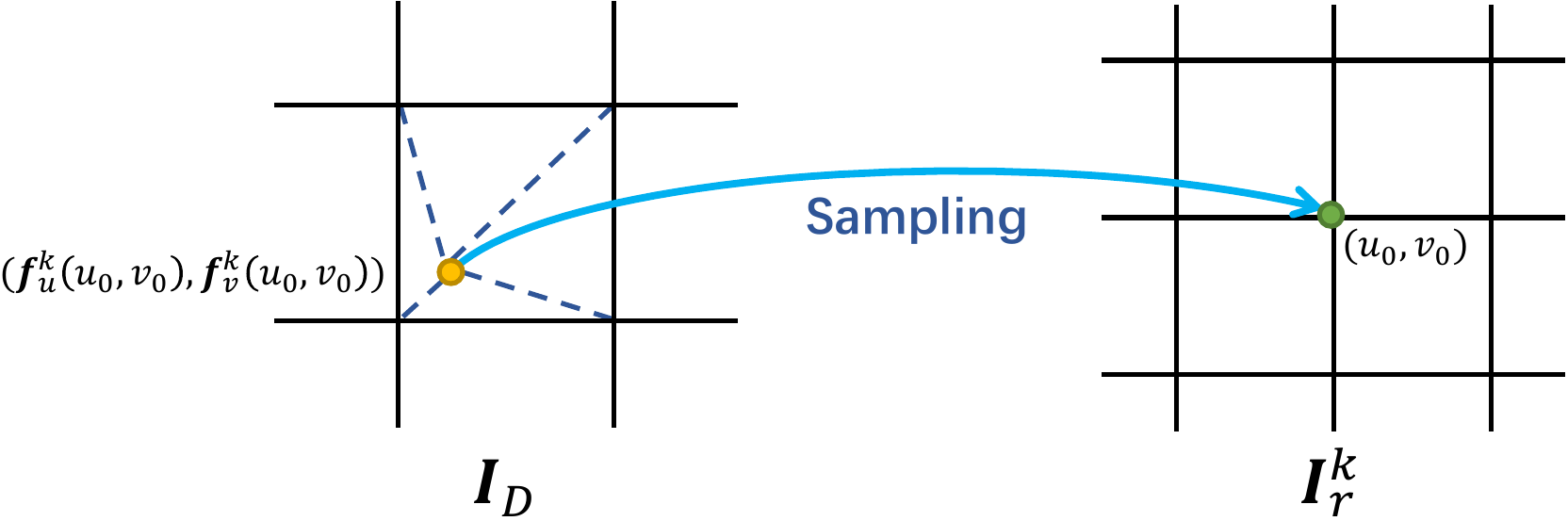}
	\end{center}
	\vspace{-0.16in}
	\caption{Visualization of the rectification process of a certain pixel on the rectified image based on the warping flow. $\bm{I}_D$ and $\bm{I}_r^{k}$ are the input distorted image and output rectified image, respectively.}
	\label{fig:sampling}
	\vspace{-0.1in}
\end{figure}

\vspace{-0.05in}
\subsection{Progressive Rectification Module}
Given the background-excluded image,
the progressive rectification module progressively corrects it toward a distortion-free one.
Specifically, we design a compact recurrent architecture to refine the rectification result estimated at the previous iteration. 
Through iterative refinements, 
the distortions in the input distorted document images are progressively corrected and finally converge to a relatively steady and accurate status. 

As shown in Fig.~\ref{fig:2}, given the background-excluded image $\bm{I}_d \in \mathbb{R}^{H \times W \times 3}$ obtained by the document localization module, we estimate the warping flow iteratively and get the sequence $\{\bm{f}^1,\cdots,\bm{f}^K\}$, where $\bm{f}^{k}=(\bm{f}_u^{k}, \bm{f}_v^{k})$ is the predicted warping flow at the $k^{th}$ iteration, and $K$ is the total iteration number.
Note that the two channel of the warping flow $\bm{f}^{k} \in \mathbb{R}^{H \times W \times 2}$ denote the horizontal and the vertical coordinate mapping (\emph{i.e.}, $\bm{f}_u^{k}$ and $\bm{f}_v^{k}$), respectively. With $\bm{f}^k$ predicted at the $k^{th}$ iteration, as illustrated in Fig.~\ref{fig:sampling}, the rectified image $\bm{I}_r^{k}$ can be obtained by the warping operation based on the bilinear sampling as follows,
\begin{equation}\label{equ:task}
	\bm{I}_r^{k}(u_0,v_0) = \bm{I}_D(\bm{f}_u^{k}(u_0,v_0), \bm{f}_v^{k}(u_0,v_0)),
\end{equation}
where $(u_{0},v_{0})$ is the integer pixel coordinate in rectified image, and $(\bm{f}_u^{k}(u_0,v_0), \bm{f}_v^{k}(u_0,v_0))$ is the predicted decimal pixel coordinate in distorted image. 

For convenience of understanding, we divide the progressive rectification module into three blocks, including (1) distorted feature encoder, (2) rectified feature generator, and (3) warping flow updater. In the following, we separately detail the three blocks.

\setlength{\parskip}{0.5em}
\noindent
\textbf{Distorted feature encoder.} Given the input image $\bm{I}_d \in \mathbb{R}^{H\times W\times3}$, we use a convolutional network $E_{\theta}$ to extract features from distorted image $\bm{I}_d$. $E_{\theta}$ consists of 6 residual blocks~\cite{he2016deep} and stride the feature maps every two blocks, followed by two parallel convolutional layers. The two parallel layers produce features $\bm{c}_0 \in \mathbb{R}^{\frac{H}{8} \times \frac{W}{8} \times D}$ and $\bm{h}_0 \in \mathbb{R}^{\frac{H}{8} \times \frac{W}{8} \times D}$, respectively, where we set channel dimension $D=128$. $\bm{c}_0$ denotes the distorted features, and $\bm{h}_0$ serves as the initial hidden state for warping flow updater. Note that both $\bm{c}_0$ and $\bm{h}_0$ need to be calculated only once.

\begin{figure}[t]
	\begin{center}
		\includegraphics[width=0.97\linewidth]{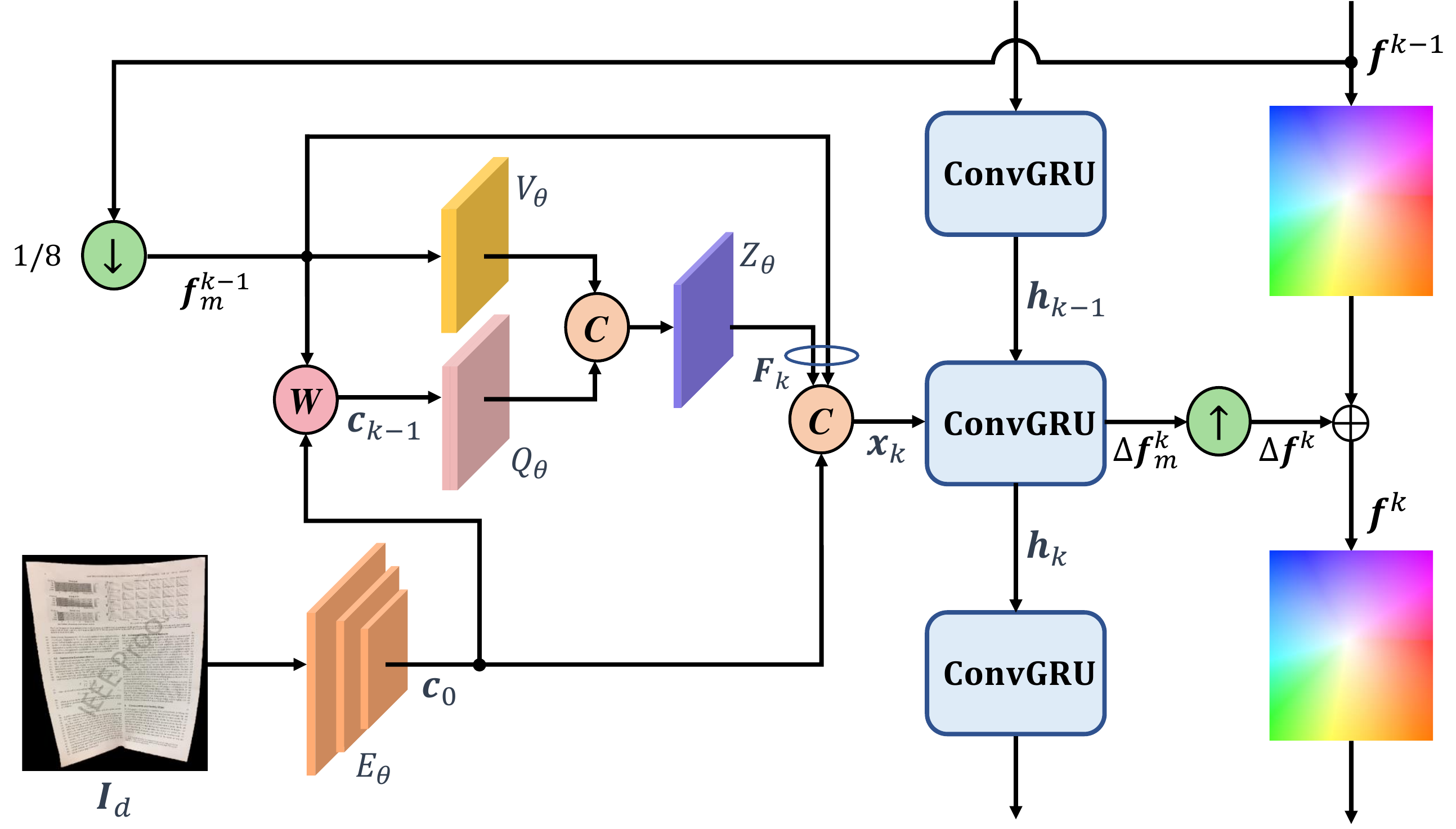}
	\end{center}
	\vspace{-0.05in}
	\caption{Illustration of the warping flow estimation at the $k^{th}$ iteration. Given the distorted features $\bm{c}_0$ and predicted warping flow $\bm{f}^{k-1}$, it outputs the current warping flow $\bm{f}^k$. $\bm{W}$ represents the bilinear sampling operation of warping. ``$\bm{C}$" and ``$\bm{+}$" denote concatenation over channel and element-wise addition, respectively. ``$\downarrow$" and ``$\uparrow$" denote the bilinear downsampling and the learnable upsampling module, respectively.}
	\label{fig:4}
	\vspace{-0.1in}
\end{figure}

\setlength{\parskip}{0.5em}
\noindent
\textbf{Rectified feature generator.} As shown in Fig.~\ref{fig:4}, we take the $k^{th}$ iteration as an example for illustration. Given the distorted features $\bm{c}_0$ from the distorted feature encoder and the warping flow $\bm{f}^{k-1}$ predicted at the $(k-1)^{th}$ iteration, we first downsample $\bm{f}^{k-1}$  and get the warping flow $\bm{f}_{m}^{k-1}=(\bm{f}_{mu}^{k-1}, \bm{f}_{mv}^{k-1})$ at 1/8 resolution. Then, we unwarp the feature maps $\bm{c}_0$ toward rectified domain using predicted $\bm{f}_{m}^{k-1}$ based on bilinear sampling, and obtain features $\bm{c}_{k-1}$ as follows,
\setlength{\parskip}{0em} 
\begin{equation}
	\bm{c}_{k-1}(x,y) = \bm{c}_0(\bm{f}^{k-1}_{mu}(x,y), \bm{f}^{k-1}_{mv}(x,y)),
\end{equation}
where $(x,y)$ is the integer pixel coordinate in $\bm{c}_{k-1}$, and $(\bm{f}_{mu}^{k-1}(x,y), \bm{f}_{mv}^{k-1}(x,y))$ is the predicted decimal pixel coordinate in $\bm{c}_0$. Note that the initial warping flow $\bm{f}^{0} \in \mathbb{R}^{H \times W \times 2}$ is initialized as the coordinate map of the pixels in $\bm{I}_d$. In addition, the warping operation is implemented based on bilinear interpolation. Therefore, we can compute the gradients to the input feature map $\bm{c}_0$ and warping flow $\bm{f}_{m}^{k-1}$ for backpropagation, according to classic STN~\cite{jaderberg2015spatial}, and the module can be trained in an end-to-end manner. Then the warped feature map $\bm{c}_{k-1}$ is processed by a convolutional module $Q_\theta$ which consists of two convolutional layers, and produce features $Q_\theta (\bm{c}_{k-1}) \in \mathbb{R}^{\frac{H}{8} \times \frac{W}{8} \times D_q}$, where we set $D_q=192$. 

\setlength{\tabcolsep}{5mm}
\begin{table}[t]
	\small
	\centering
	\caption{The structure of the distorted feature encoder and rectified feature generator in DocScanner-B.}
	\vspace{-0.1in}
	\begin{tabular}{c|c|c}  
		\Xhline{1.5\arrayrulewidth}
		Layer & Output size & Operation\\   
		\hline
		$E_{\theta}$ & 256 $\times$ 36 $\times$ 36 & $\begin{matrix} 7\times7, 64, stride~2 \\ 3\times3, 64, stride~1 \\ 3\times3, 64, stride~1 \\ 3\times3, 96, stride~1\\ 3\times3, 96, stride~2 \\ 3\times3, 128, stride~1\\ 3\times3, 128, stride~2 \\ 1\times1, 256, stride~1 \end{matrix}$\\
		
		\hline
		$V_{\theta}$ & 64 $\times$ 36 $\times$ 36 & $\begin{matrix} 7\times7, 128, stride~1 \\ 3\times3, 64, stride~1  \end{matrix}$ \\
		
		\hline
		$Q_{\theta}$ & 192 $\times$ 36 $\times$ 36 & $\begin{matrix} 1\times1, 224, stride~1 \\ 3\times3, 192, stride~1  \end{matrix}$ \\
		
		\hline
		$Z_{\theta}$ & 128 $\times$ 36 $\times$ 36 & $\begin{matrix} 3\times3, 128, stride~1  \end{matrix}$ \\
		\Xhline{1.5\arrayrulewidth}	
	\end{tabular}
	\vspace{-0.1in}
\end{table}

Additionally, another convolutional module $V_\theta$ that consists of two convolutional layers is used to extract features from the predicted warping flow $\bm{f}_m^{k-1}$, and output features $V_\theta (\bm{f}_m^{k-1}) \in \mathbb{R}^{\frac{H}{8} \times \frac{W}{8} \times D_v}$, where we set $D_v=64$. Then, we concatenate $Q_\theta (\bm{c}_{k-1})$ and $V_\theta (\bm{f}_m^{k-1})$ along the channel dimension into a single feature map, which is fused by a following convolutional layer $Z_\theta$. Finally, we concatenate the output features and the downsampled warping flow $\bm{f}_m^{k-1}$ to generate the rectified feature map $\bm{F}_k \in \mathbb{R}^{\frac{H}{8} \times \frac{W}{8} \times D}$.
It carries the content and the structural information of the current rectified image estimated at the previous iteration, which is differentiated and processed by the following updater to estimate a further refinement.

\begin{figure}[t]
	\begin{center}
		\includegraphics[width=1\linewidth]{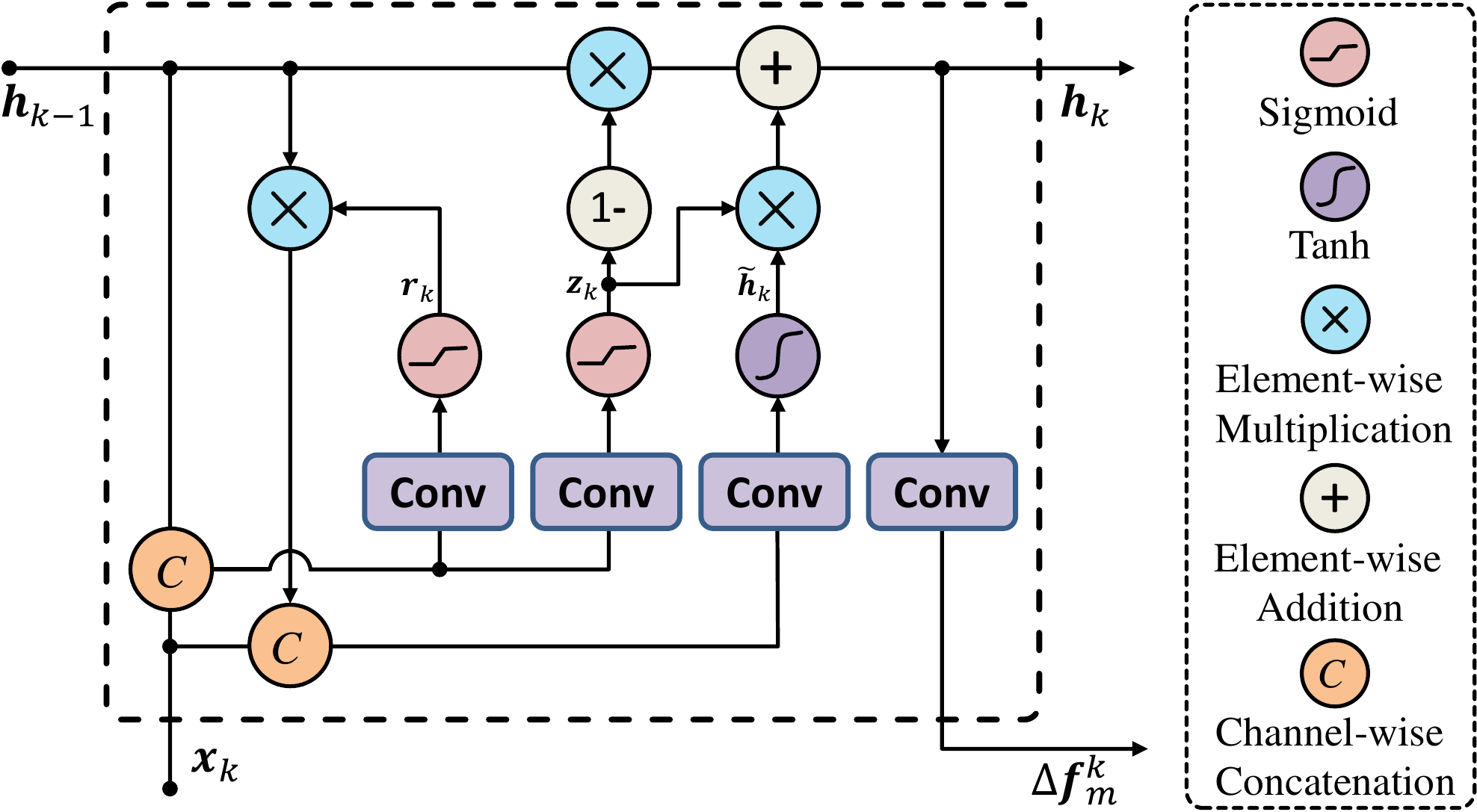}
	\end{center}
	\caption{Inner structure of the ConvGRU, a modified version of GRU~\cite{cho2014properties}.}
	\label{fig:gru}
	\vspace{-0.1in}
\end{figure}

\setlength{\parskip}{0.5em}
\noindent
\textbf{Warping flow updater.} As shown in Fig.~\ref{fig:4}, we concatenate distorted features $\bm{c}_0$ and the rectified features $\bm{F}_k$ along the channel dimension into a single feature map $\bm{x}_{k} \in \mathbb{R}^{\frac{H}{8} \times \frac{W}{8} \times 2D}$, which serves as the input of the recurrent unit at the $k^{th}$ iteration. We use a convolution-based gated recurrent unit (GRU) as the recurrent unit as many other tasks~\cite{tokmakov2017learning,teed2020raft,zhou2021r}. As shown in Fig.~\ref{fig:gru}, it is a variant of GRU~\cite{cho2014properties}, in which the fully connected layers are replaced by the convolutional layers. For the $k^{th}$ iteration, it processes the input features $\bm{x}_{k} \in \mathbb{R}^{\frac{H}{8} \times \frac{W}{8} \times 2D}$ and the hidden state $\bm{h}_{k-1} \in \mathbb{R}^{\frac{H}{8} \times \frac{W}{8} \times D}$, and outputs the hidden states $\bm{h}_k \in \mathbb{R}^{\frac{H}{8} \times \frac{W}{8} \times D}$ as follows,
\begin{gather}
	\bm{z}_k=\sigma(Conv_{3\times3}([\bm{h}_{k-1},\bm{x}_k],\bm{W}_z)), \\
	\bm{r}_k=\sigma(Conv_{3\times3}([\bm{h}_{k-1},\bm{x}_k],\bm{W}_r)), \\
	\widetilde{\bm{h}}_k=tanh(Conv_{3\times3}([\bm{r}_k\odot \bm{h}_{k-1},\bm{x}_k],\bm{W}_h)),  \\
	\bm{h}_k=(1-\bm{z}_k)\odot \bm{h}_{k-1}+\bm{z}_k\odot \widetilde{\bm{h}}_k, 
\end{gather}
where $\sigma$ and $\odot$ represent Sigmoid function and element-wise multiplication operation, respectively.
Followed by $\bm{h}_k$ is two convolutional layers that produce the residual displacement $\Delta \bm{f}^k_m \in \mathbb{R}^{\frac{H}{8} \times \frac{W}{8} \times 2}$. 

\setlength{\parskip}{0em}
To upsample the obtained $1/8$ scale $\Delta \bm{f}^k_m$ to full resolution ($H \times W$), we introduce a learnable upsampling module~\cite{feng2021doctr}.
Specifically, we first exploit two convolutional layers (stride 1) to process the hidden state $\bm{h}_k \in \mathbb{R}^{\frac{H}{8} \times \frac{W}{8} \times D}$, and reshape the output to a $\frac{H}{8} \times \frac{W}{8} \times 8\times8\times9$ map.
Then, we perform softmax on the last dimension of it and get the weight matrix.
Next, using the obtained weight matrix, we take a weighted combination over the $3\times3$ neighborhood of each pixel in $\Delta \bm{f}^k_m$.
Finally, the obtained $\frac{H}{8} \times \frac{W}{8} \times 8\times8 \times 2$ map is permuted and reshaped to the full resolution residual displacement map $\Delta \bm{f}^k \in \mathbb{R}^{H \times W \times 2}$.

After that, $\Delta \bm{f}^k$ is used to update the current warping flow $\bm{f}^k$ as follows,
\begin{equation}
	\bm{f}^k = \bm{f}^{k-1} + \Delta \bm{f}^k. \label{update}
\end{equation}
As shown in Fig.~\ref{fig:2}, after $K$ iterations, based on Equation~\eqref{equ:task}, we obtain the rectified image $\bm{I}_r^{K}$ by warping the distorted image $\bm{I}_D$ with the final predicted $\bm{f}^K$.

\subsection{Training Loss Function}
During the training of the progressive rectification module, the loss is calculated over all $K$ iterations as follows,
\begin{equation}\label{equ:loss_total}
	\mathcal{L}=\sum_{k=1}^K \lambda ^{K-k} \mathcal{L}^{(k)}, 
\end{equation}
where $\lambda ^{K-k}$ is the weight of the $k^{th}$ iteration which increases exponentially ($\lambda < 1$). At the $k^{th}$ iteration, the loss is defined as the weighted summation of the warping flow regression loss $\mathcal{L}_{f}^{(k)}$ and the proposed circle-consistency loss $\mathcal{L}_{line}^{(k)}$ as follows,
\begin{equation}\label{equ:lossfun}
	\mathcal{L}^{(k)}=\mathcal{L}_{f}^{(k)} + \alpha  	\mathcal{L}_{line}^{(k)},
\end{equation}
where $\alpha$ is a constant weighting factor.
$\mathcal{L}_{f}^{(k)}$ is defined as the $L_1$ distance between the predicted warping flow $\bm{f}^k$ and its given ground truth $\bm{f}_{gt}$ as follows,
\begin{equation}
	\mathcal{L}_{f}^{(k)} = \left \| \bm{f}_{gt} - \bm{f}^k \right \|_1.
\end{equation}
The proposed circle-consistency loss $\mathcal{L}_{line}^{(k)}$ works as a regularizer, which imposes straight-line constraint along rows and columns in rectified image. We detail it in the following.

\setlength{\parskip}{0em}
\subsubsection{Circle-consistency Loss}
Equation~\eqref{equ:task} shows that during the rectification process, the pixel in rectified image $\bm{I}_r^{k}$ is filled with the corresponding pixel sampled in distorted image $\bm{I}_D$ based on the predicted warping flow $\bm{f}^k$. This predicted warping flow $\bm{f}^k$ is termed as backward warping flow. We further introduce forward warping flow $\bm{g}=(\bm{g}_x,\bm{g}_y)$ from the dataset, which maps the pixel $(x_0,y_0)$ in distorted image $\bm{I}_D$ to pixel $(\bm{g}_x(x_0,y_0),\bm{g}_y(x_0,y_0))$ in rectified image $\bm{I}_r^{k}$ as follows,
\begin{equation}
	\bm{I}_r^{k}(\bm{g}_x(x_0,y_0),\bm{g}_y(x_0,y_0)) = \bm{I}_D(x_0, y_0),
\end{equation}
where $(x_{0},y_{0})$ is the integer pixel coordinate in the distorted image $\bm{I}_D$, while $(\bm{g}_x(x_0,y_0), \bm{g}_y(x_0,y_0))$ is the corresponding decimal pixel coordinate in the rectified image $\bm{I}_r^{k}$.

\begin{figure}[t]
	\begin{center}
		\includegraphics[width=0.98\linewidth]{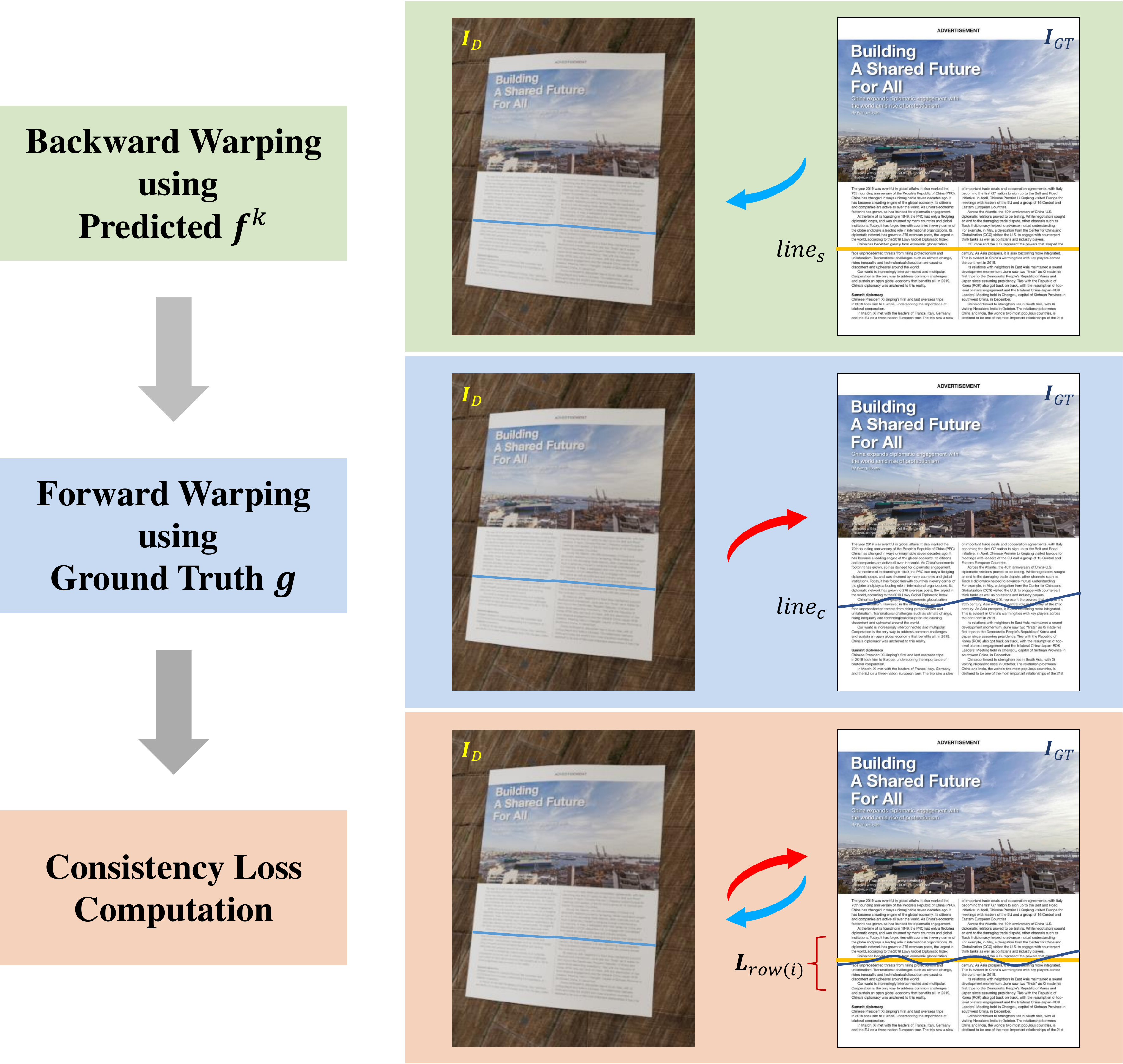}
	\end{center}
	\vspace{-0.04in}
	\caption{Illustration of the circle-consistency loss. After warping a line pixels $line_s$ using the predicted backward warping flow $\bm{f}^k$ and ground truth forward warping flow $\bm{g}$, the output line $line_c$ should be consistent with itself $line_s$ under perfect prediction. Based on this observation, the circle-consistency loss is defined by computing the distortion of $line_c$.}
	\label{fig:lineloss}
	\vspace{-0.08in}
\end{figure}

We propose circle-consistency loss, based on the circle-consistency introduced by the backward warping and the forward warping operations. It consists of two terms, along the row and column direction respectively. As shown in Fig.~\ref{fig:lineloss}, we take the row distortion term as an example. Specifically, we first map the pixels of $i^{th}$ row (\emph{i.e.}, $line_s$) in ground truth document image to $\bm{I}_D$, based on the predicted backward warping flow $\bm{f^{k}}$. Secondly, we map these pixels back to the ground truth document image again, using the ground truth forward warping flow $\bm{g}$. After the above two steps, we get a curved line $line_c$, which shall be the straight line $line_s$ when the backward warping flow in the first step is perfectly estimated. Hence, we define the $i^{th}$ row circle-consistency loss $\mathcal{L}_{row(i)}$ as the deviation of row coordinate of the estimated curved line $line_c$ as follows,
\begin{equation}
	\mathcal{L}_{row(i)}=\frac{1}{W}\sum_{k=1}^W {\left \| x{(i,k)} - \overline{x_i} \right \|}_2^2,
\end{equation} 
where $\overline{x_i}$ denotes the averaged row coordinate of the curved line $line_c$, and $x(i,k)$ denotes the row coordinate of the $k^{th}$ pixel of $line_c$. $\mathcal{L}_{row(i)}$ measures the distortion of the $i^{th}$ row which should be zero in case of perfect rectification.

Similarly, we can calculate the column circle-consistency term $\mathcal{L}_{col(j)}$ for the $j^{th}$ column.
Then, the total circle-consistency loss $\mathcal{L}_{line}$ is calculated over all rows and columns as follows,
\begin{equation}
	\mathcal{L}_{line}=\frac{1}{W}\sum_{i=1}^W \mathcal{L}_{row(i)} + \frac{1}{H}\sum_{j=1}^H \mathcal{L}_{col(j)},
\end{equation}
where $(H,W)$ is the shape of the predicted warping flow $\bm{f}^{k}$.

\section{Experiments}
\subsection{Datasets}
We train our DocScanner on the Doc3D dataset~\cite{9010747} and evaluate it on the DocUNet Benchmark dataset~\cite{8578592}. In the following, we elaborate the two datasets respectively.

\smallskip
\noindent
\textbf{Doc3D.} Doc3D dataset~\cite{9010747} is the largest dataset to date for document image rectification. It is created by the real document data and rendering software, \emph{i.e.}, Blender\footnote{https://www.blender.org/}. The dataset consists of 100k distorted document images. For each distorted image, there are corresponding ground truth 3D coordinate map, albedo map, normals map, depth map, forward warping flow map, and backward warping flow map.

\smallskip
\noindent
\textbf{DocUNet Benchmark.} The challenging DocUNet Benchmark dataset~\cite{8578592} is a widely-used dataset for document image rectification. It comprises 130 photos of real paper documents captured by mobile cameras. The documents include various types such as receipts, letters, fliers, magazines, academic papers, and books, \emph{etc}. Besides, their distortion and background are various to cover different levels of difficulty.

Notably, we observe that the $127^{th}$ and $128^{th}$ distorted document images are rotated by 180 degrees,
which do not match the ground truth documents.
This inconsistency is ignored by existing methods~\cite{8578592,9010747, liu2020geometric, xie2020dewarping, das2021end, feng2021doctr,DDCPXIE, zhang2022marior,jiang2022revisiting, ma2022learning}.
In our experiments,
we use the corrected dataset.

\setlength{\parskip}{0em}
\subsection{Evaluation Metrics}
We use three evaluation schemes to quantitatively evaluate the performance of DocScanner in terms of (a) rectification distortion, (b) Optical Character Recognition (OCR) accuracy, and (c) image similarity. Firstly, for rectification distortion, we use Local Distortion (LD)~\cite{7866848} as recommended in~\cite{8578592,9010747, liu2020geometric, xie2020dewarping, das2021end, feng2021doctr, DDCPXIE,zhang2022marior,jiang2022revisiting, ma2022learning}. Moreover, we propose a new metric, namely Line Distortion (Li-D), to further evaluate the global distortion of the rectified document images. Secondly, for OCR accuracy, we choose Edit Distance (ED)~\cite{levenshtein1966binary} and Character Error Rate (CER)~\cite{morris2004and} to evaluate the utility of our method on text recognition, following~\cite{8578592,9010747,das2021end, feng2021doctr,zhang2022marior,jiang2022revisiting,ma2022learning}. Thirdly, for image similarity, we use Multi-Scale Structural Similarity (MS-SSIM)~\cite{1292216} as previous works~\cite{8578592,9010747,liu2020geometric,das2020intrinsic, xie2020dewarping,das2021end, feng2021doctr, DDCPXIE,zhang2022marior,jiang2022revisiting,ma2022learning} suggest.

\setlength{\tabcolsep}{0.15mm}
\begin{table*}[t]
	\caption{Quantitative comparisons of the existing learning-based rectification methods in terms of image similarity, distortion metrics, OCR performance, and running efficiency on the DocUNet Benchmark dataset~\cite{8578592}. ``$\uparrow$'' indicates the higher the better and ``$\downarrow$'' means the opposite.}
	\vspace{-0.05in}
	\centering
	
	\begin{tabular}{l|c|c|cc|cc|cc}  
		
		\Xhline{2\arrayrulewidth}
		\textbf{Methods} & \textbf{Venue} &\textbf{MS-SSIM} $\uparrow$ &\textbf{LD} $\downarrow$ &\textbf{Li-D} $\downarrow$ &\textbf{ED} $\downarrow$ &\textbf{CER} $\downarrow$  &\textbf{FPS} $\uparrow$ 
		&\textbf{Para.(M)} \\  
		
		\hline\hline
		
		Distorted & - & 0.2459 & 20.51 & 5.66 & 2111.56/1552.22 & 0.5352/0.5089  & - & - \\ 
		
		\hline
		
		DocUNet~\cite{8578592} & \emph{CVPR'18} & 0.4103 & 14.19 & 3.19 & 1933.66/1259.83 & 0.4632/0.3966  &  & 58.6 \\
		
		AGUN~\cite{liu2020geometric} & \emph{PR'18} & - & -  & - & - & -  & - & - \\
		
		DocProj~\cite{li2019document} & \emph{TOG'19}  & 0.2946 & 18.01 & 5.00 & 1712.48/1165.93 & 0.4267/0.3818 & - & 47.8 \\ 
		
		
		FCN-based~\cite{xie2020dewarping} & \emph{DAS'20} & 0.4477 & 7.84 & 2.04 & 1792.60/1031.40 & 0.4213/0.3156  & 1.49 & 23.6 \\  
		
		DewarpNet~\cite{9010747} & \emph{ICCV'19} & 0.4735 & 8.39 & 2.31 & 885.90/525.45 & 0.2373/0.2102  & 8.17 & 86.9 \\
		
		PWUNet~\cite{das2021end} & \emph{ICCV'21} & 0.4915 & 8.64 & 2.34  & 1069.28/743.32 & 0.2677/0.2623  & - & - \\    
		
		DocTr~\cite{feng2021doctr} & \emph{MM'21} & 0.5105 & 7.76 & 2.11 & 724.84/464.83 & 0.1832/0.1746  & 7.74 & 26.9 \\
		
		DDCP~\cite{DDCPXIE} & \emph{ICDAR'21} & 0.4729 & 8.99 & 2.20 & 1442.84/745.35 & 0.3633/0.2626 & \textcolor{red}{\textbf{14.09}} & 13.3 \\

		DocGeoNet~\cite{feng2022docgeonet} & \emph{ECCV'22} & 0.5040 & 7.71 & 2.22 & 713.94/\textcolor{blue}{\textbf{379.00}} & 0.1821/\textcolor{blue}{\textbf{0.1509}}  & 8.57 & 24.8 \\ 

        Marior~\cite{zhang2022marior} & MM'22 & 0.4780 & \textcolor{red}{\textbf{7.44}} & 2.03 & 776.22/593.80 & 0.1928/0.2136  & - & -\\        
    
		RDGR~\cite{jiang2022revisiting} &  \emph{CVPR'22} & 0.4968 & 8.51 & 2.12 & 729.52/420.25 & \textcolor{blue}{\textbf{0.1717}}/0.1559 &- &- \\

            PaperEdge~\cite{ma2022learning} &  \emph{SIGGRAPH'22} & 0.4724 & 7.99 & \textcolor{red}{\textbf{1.83}} & 777.76/\textcolor{red}{\textbf{375.60}} & 0.2014/0.1541 & \textcolor{blue}{\textbf{13.95}} & 36.6 \\

		\hline 
		
		DocScanner-T & - & 0.5123 & 7.92 & 2.04 & 809.46/501.82 & 0.2068/0.1823 & 10.81 & \textcolor{red}{\textbf{2.6}} \\ 
		
		DocScanner-B & - & \textcolor{red}{\textbf{0.5134}} & 7.62 & 1.88 & \textcolor{blue}{\textbf{671.48}}/434.11 & 0.1789/0.1652  & 10.03 & \textcolor{blue}{\textbf{5.2}} \\ 
		
		DocScanner-L & - & \textcolor{red}{\textbf{0.5178}} & \textcolor{blue}{\textbf{7.45}} & \textcolor{blue}{\textbf{1.86}} & \textcolor{red}{\textbf{632.34}}/390.43 & \textcolor{red}{\textbf{0.1648}}/\textcolor{red}{\textbf{0.1486}}   & 9.52 & 8.5 \\ 
		
		\Xhline{2\arrayrulewidth}	
	\end{tabular}
	\label{t1}
\end{table*}

\smallskip
\noindent
\textbf{Local Distortion.} Local distortion (LD)~\cite{7866848} first registers the rectified image with the ground truth one using a dense SIFT-flow~\cite{5551153} $(\Delta \bm{x}, \Delta \bm{y})$, where $\Delta \bm{x}$ and $\Delta \bm{y}$ denote the horizontal and vertical displacement map of the matched pixels from the ground truth image to the rectified one, respectively. 
 Then, LD is calculated as the mean value of the $L_2$ distance among all matched pixels, which measures the average local deformation of the rectified image. Note that, for a fair comparison, all the rectified images and the ground truth images are resized to a 598,400-pixel area, as suggested in~\cite{8578592,9010747,liu2020geometric,xie2020dewarping,das2021end, feng2021doctr, DDCPXIE,zhang2022marior,jiang2022revisiting,ma2022learning}.

\smallskip
\noindent
\textbf{Line Distortion.}
We propose Line Distortion (Li-D) as a supplementary metric to further evaluate the global distortion of the rectified images. Specifically, the dense SIFT-flow~\cite{5551153} $(\Delta \bm{x}, \Delta \bm{y})$ from the ground truth scanned image to the rectified one is first computed. Then, we calculate the standard deviation of all column vectors in the $\Delta \bm{x}$ and all row vectors in the $\Delta \bm{y}$, which measure the deformation of a certain rectified row and column pixels, respectively. Finally, we take the mean of all the standard deviation values to obtain the overall Line Distortion (Li-D) value. 

Compared to the typical metric Local Distortion (LD)~\cite{7866848}, the proposed Line Distortion (Li-D) computes the average deformation of the row and column pixels. In another word, Li-D focuses more on the global distortions. The less distortion of the rectified image, the lower the value.
Note that if there is only global misalignment (i.e. scaling and translation) between two images, the Li-D should be 0 but such global misalignments are not considered for this task.

\smallskip
\noindent
\textbf{ED and CER.} Edit Distance (ED)~\cite{levenshtein1966binary} quantifies how dissimilar two strings are to one another. It is defined based on the minimum number of operations required to transform one string into the reference one, which can be efficiently calculated using the dynamic programming algorithm. Specifically, the involved operations include deletions $(d)$, insertions $(i)$, and substitutions $(s)$. Then, Character Error Rate (CER) can be computed as follows,
\begin{equation}\label{equ:cer}
CER=(d+i+s)/{N_c} ,
\end{equation}
where $N_c$ is the character number of the reference string. It represents the percentage of characters in the reference text that was incorrectly recognized in the distorted image. The lower the CER value (with 0 being a perfect score), the better the performance of the rectification method. We use Tesseract (v5.0.1)~\cite{4376991} as the OCR engine to recognize the text string of the rectified image and the ground truth image, as recommended in previous works~\cite{8578592,9010747,liu2020geometric,das2020intrinsic, xie2020dewarping,das2021end, feng2021doctr, DDCPXIE,zhang2022marior,jiang2022revisiting,ma2022learning}.

\smallskip
\noindent
\textbf{MS-SSIM.} The Structural SIMilarity (SSIM)~\cite{1284395} measures the similarity of mean value and variance within each image patch between two images. Considering that the perceivability of image details depends on the sampling density of the image, Multi-Scale Structural Similarity (MS-SSIM)~\cite{1292216} builds a Gaussian pyramid for the rectified image and the corresponding ground truth image, respectively. Then, MS-SSIM is calculated as the weighted summation of SSIM~\cite{1284395}  across multiple scales. Specifically, all the rectified and ground truth flatbed-scanned images are first resized to a 598,400-pixel area, as recommended in DocUNet~\cite{8578592}. Then, we build a 5-level-pyramid for MS-SSIM and the weight for each level is set as 0.0448, 0.2856, 0.3001, 0.2363, 0.1333, which is inherited from the original implementation of MS-SSIM~\cite{1292216}.

\subsection{Training Details}
The whole framework of DocScanner is implemented in Pytorch~\cite{paszke2017automatic}. We train the document localization module and progressive rectification module independently on the Doc3D dataset~\cite{9010747}. We detail their training in the following.

\smallskip
\noindent
\textbf{Document localization module.} During training, to generalize well to real data with complex background environments, we randomly replace the background of the distorted image with the texture images from Describable Texture Dataset (DTD)~\cite{6909856}. We use Adam optimizer~\cite{kingma2014adam} with a batch size of 32. The initial learning rate is set as $1\times10^{-4}$, and reduced by a factor of 0.1 after 30 epochs. After 45 epochs, the training loss converges. The training is conducted on two NVIDIA RTX 2080 Ti GPUs. The threshold $\tau$ for binarizing the confidence map in Sec.~\ref{DLM} is empirically set as 0.5.

\smallskip
\noindent
\textbf{Progressive rectification module.} During training, we remove the background of distorted images using the ground truth masks of the foreground document regions. 
In other words, the documents are within a clean background.
To generalize well to real data with complex illumination conditions, we then add a jitter in the HSV color space to magnify illumination and document color variations. We use AdamW optimizer~\cite{loshchilov2017decoupled} with a batch size of 12. The total training iteration is set as 560k, and the learning rate reaches the maximum $1\times10^{-4}$ after 27k iterations for learning rate warm-up. We set the hyperparameters $K=12, \lambda =0.85$ (in Equation~\eqref{equ:loss_total}), $\alpha =0.5$ (in Equation~\eqref{equ:lossfun}).
Experiments are all performed on a single NVIDIA GTX 1080 Ti GPU.

\subsection{Experimental Results}
We evaluate the performance of DocScanner on the DocUNet Benchmark dataset~\cite{8578592} by quantitative and qualitative evaluation. Table~\ref{t1} shows the comparisons of our method with the existing learning-based methods on image similarity, distortion metrics, OCR accuracy, and inference efficiency. Note that for OCR accuracy evaluation, following DewarpNet~\cite{9010747} and DocTr~\cite{feng2021doctr}, we select 50 and 60 images from the DocUNet Benchmark dataset~\cite{8578592} respectively, where the text makes up the majority of content. This is because if the text is rare in an image, the character number ${N_c}$ (numerator) in Equation~\eqref{equ:cer} is a small number, leading to a large variance for CER.

For DocUNet~\cite{8578592}, DewarpNet~\cite{9010747}, FCN-based~\cite{xie2020dewarping}, DocTr~\cite{feng2021doctr}, PWUNet~\cite{das2021end}, Marior~\cite{zhang2022marior}, RDGR~\cite{jiang2022revisiting}, DocGeoNet~\cite{feng2022docgeonet}, and PaperEdge~\cite{ma2022learning}, we obtain the results based on the rectified document images of DocUNet Benchmark dataset~\cite{8578592} from the authors or the public results.
For AGUN~\cite{liu2020geometric}, there is no public official code.
Due to the two problematic samples in the DocUNet Benchmark dataset~\cite{8578592}, we can not obtain the performance.
For DocProj~\cite{li2019document} and DDCP~\cite{DDCPXIE}, we report the results based on the official code and their public pre-trained models.

\begin{figure}[t]
	\begin{center}
		\includegraphics[width=1\linewidth]{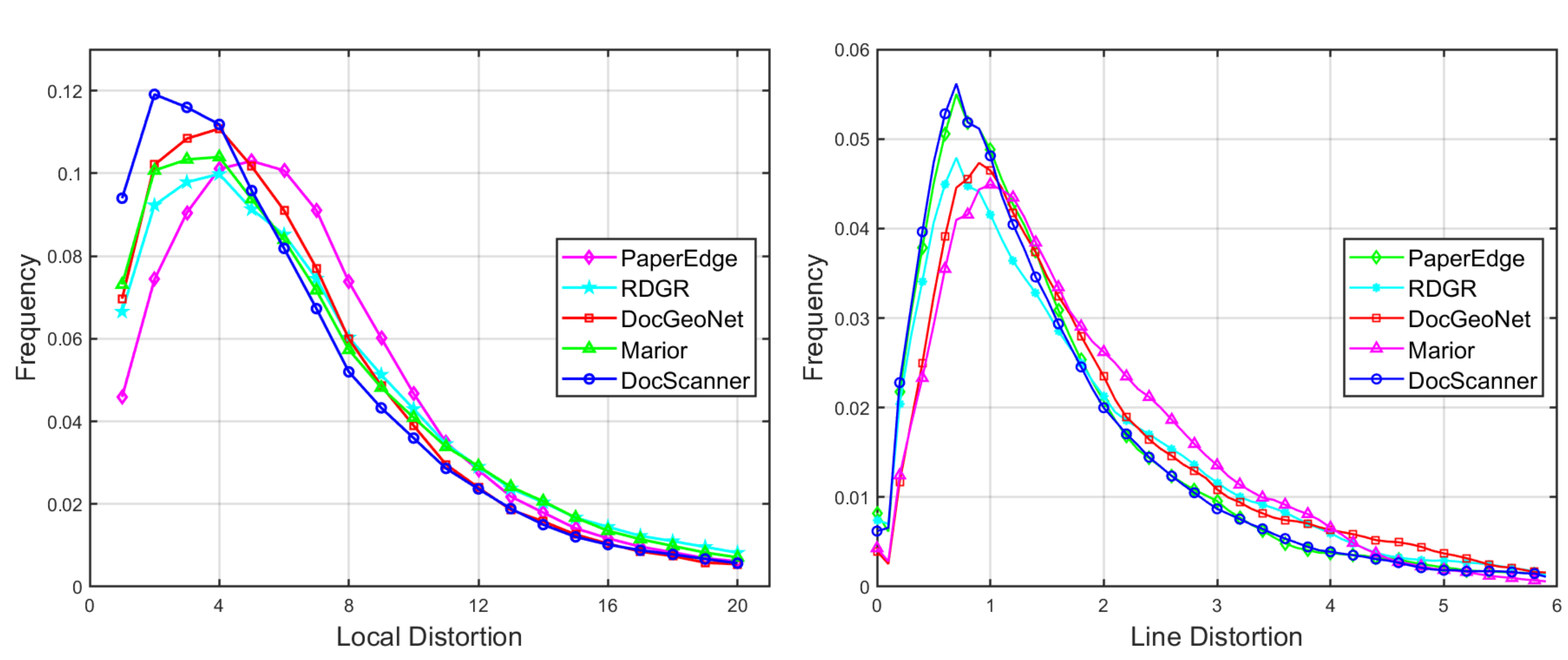}
	\end{center}
	\vspace{-0.15in}
	\caption{Comparisons of the distortion distribution curve of DocScanner-L with the state-of-the-art methods~\cite{ma2022learning,jiang2022revisiting,feng2022docgeonet,zhang2022marior}. The x-coordinate denotes the distortion extent, and the y-coordinate shows their frequency distribution among the total DocUNet Benchmark dataset~\cite{8578592}.}
	\label{fig:freq}
	\vspace{-0.05in}
\end{figure}

\begin{figure}[t]
	\begin{center}
		\includegraphics[width=1\linewidth]{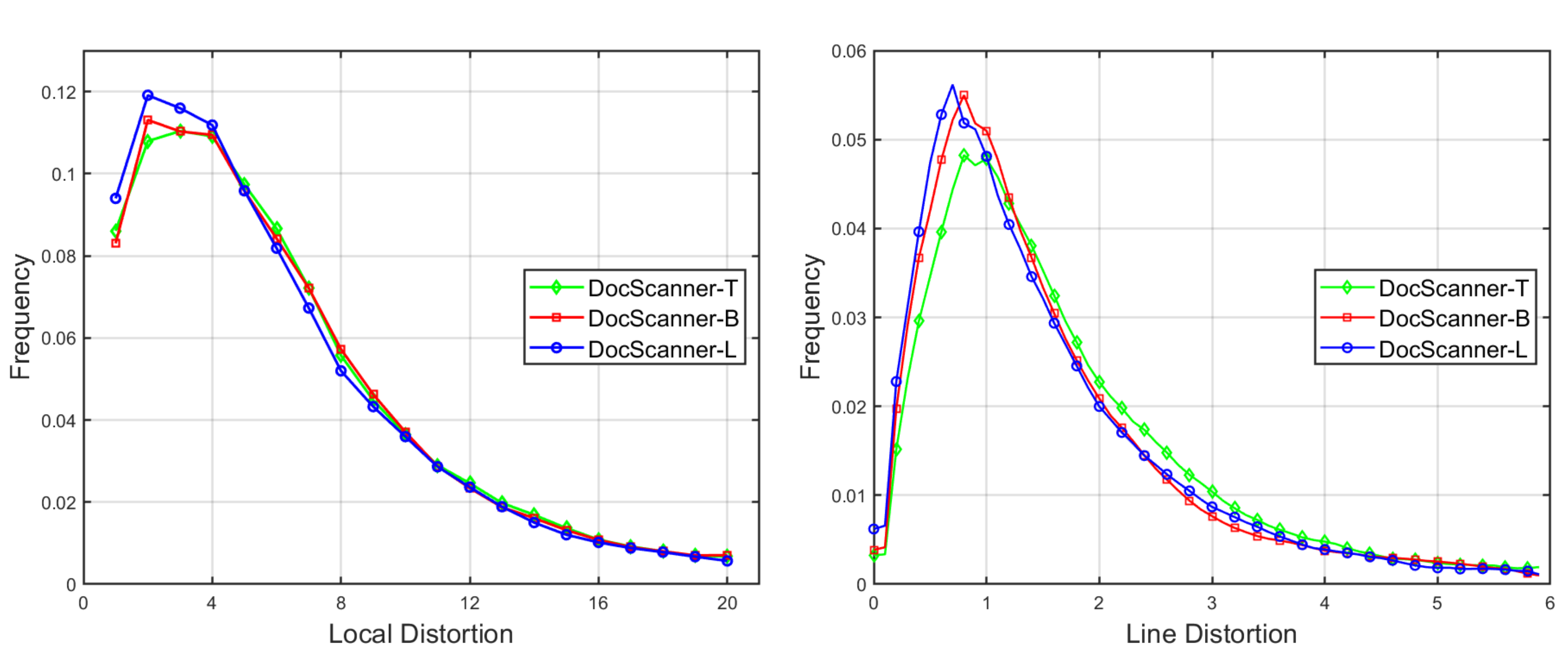}
	\end{center}
	\vspace{-0.15in}
	\caption{Comparisons of the distortion distribution curve of the three varieties of DocScanner, \emph{i.e.}, DocScanner-T, DocScanner-B, and DocScanner-L. The x-coordinate denotes the distortion extent, and the y-coordinate shows their frequency distribution among the total DocUNet Benchmark dataset~\cite{8578592}.}
	\label{fig:freq_docs}
	\vspace{-0.05in}
\end{figure}

\begin{figure*}[htbp]
	\begin{center}
		\includegraphics[width=0.98\linewidth]{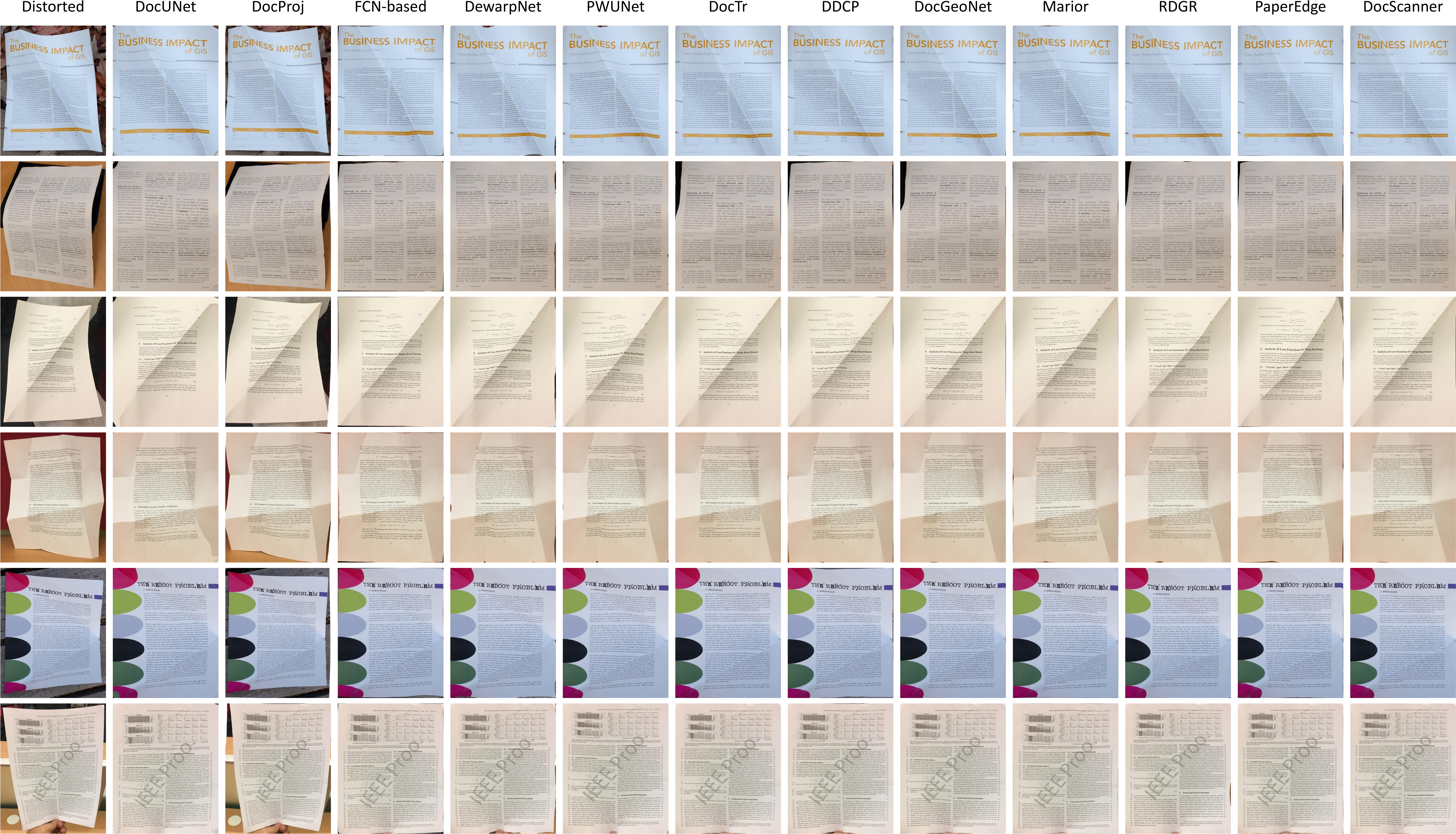}
	\end{center}
	\vspace{-0.13in}
	\caption{Qualitative comparisons with existing learning-based methods, including DocUNet~\cite{8578592}, DocProj~\cite{li2019document}, FCN-based~\cite{xie2020dewarping}, DewarpNet~\cite{9010747}, PWUNet~\cite{das2021end}, DocTr~\cite{feng2021doctr}, DDCP~\cite{DDCPXIE}, DocGeoNet~\cite{feng2022docgeonet}, Marior~\cite{zhang2022marior}, RDGR~\cite{jiang2022revisiting}, and PaperEdge~\cite{ma2022learning}.
   The rectified images of DocScanner show less distortions than the other rectification methods. Zoom in for the best view.}
	\label{fig:imgcom}
\end{figure*}

\begin{figure*}[htbp]
	\begin{center}
		\includegraphics[width=0.98\linewidth]{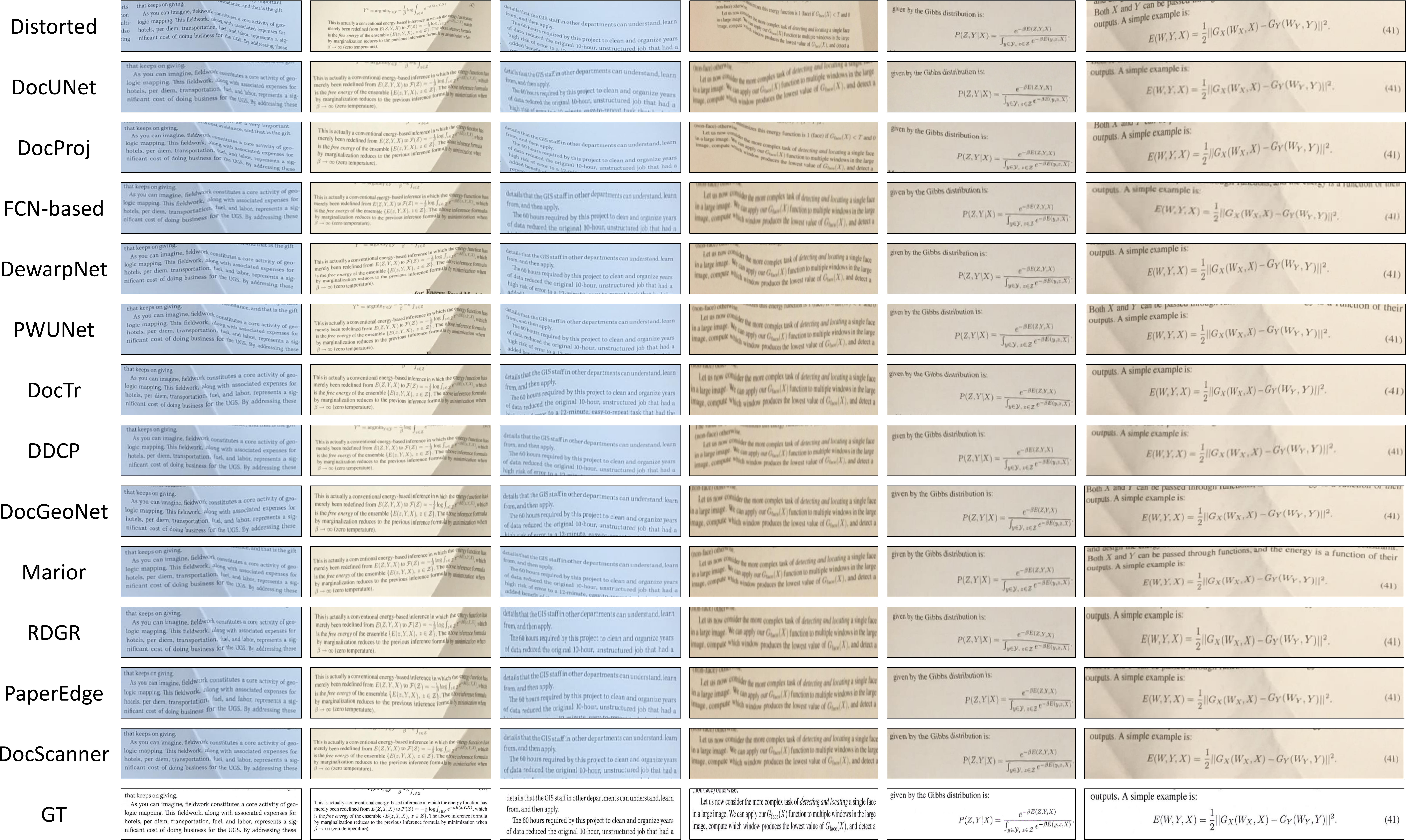}
	\end{center}
	\vspace{-0.12in}
	\caption{Qualitative comparisons of the local rectified textlines with existing learning-based methods, including DocUNet~\cite{8578592}, DocProj~\cite{li2019document}, FCN-based~\cite{xie2020dewarping}, DewarpNet~\cite{9010747}, PWUNet~\cite{das2021end}, DocTr~\cite{feng2021doctr}, DDCP~\cite{DDCPXIE}, DocGeoNet~\cite{feng2022docgeonet}, Marior~\cite{zhang2022marior}, RDGR~\cite{jiang2022revisiting}, and PaperEdge~\cite{ma2022learning}. The rectified horizontal textlines of the proposed DocScanner are much straighter than the other rectification methods.}
	\label{fig:patchcom}
\end{figure*}

\begin{figure*}[t]
	\begin{center}
		\includegraphics[width=1\linewidth]{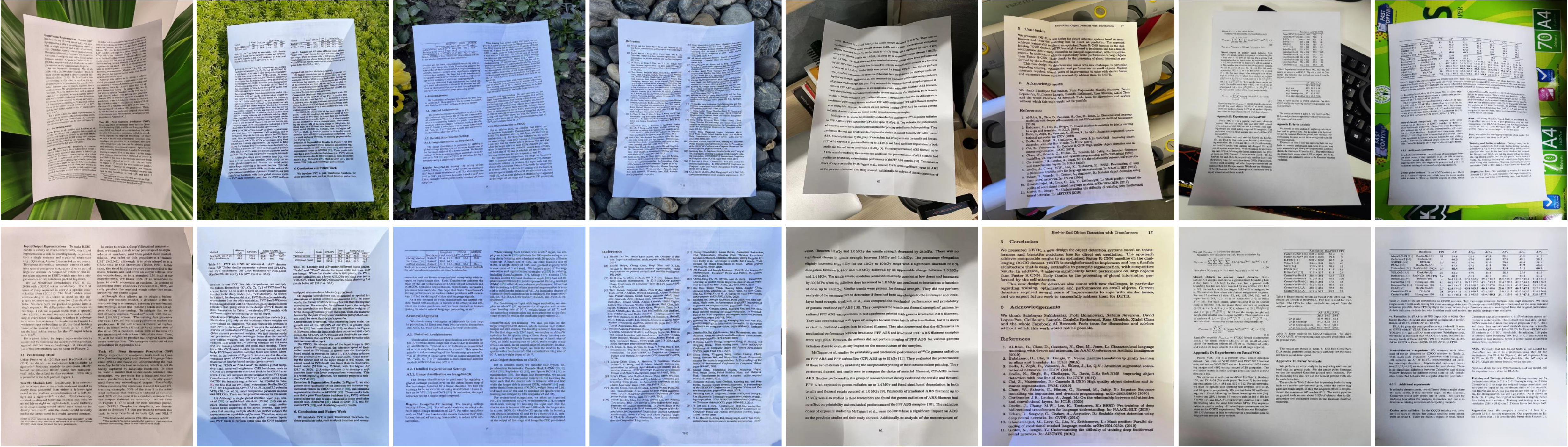}
	\end{center}
	\vspace{-0.18in}
	\caption{Robustness illustration of DocScanner on background changes. The two rows show input distorted and corresponding rectified images, respectively.
		We capture the images of deformed documents, within cluttered backgrounds under outdoor or indoor scenes during the day or night.}
\vspace{-0.08in}
\label{fig:back}
\end{figure*}

\begin{figure*}[t]
\begin{center}
	\includegraphics[width=1\linewidth]{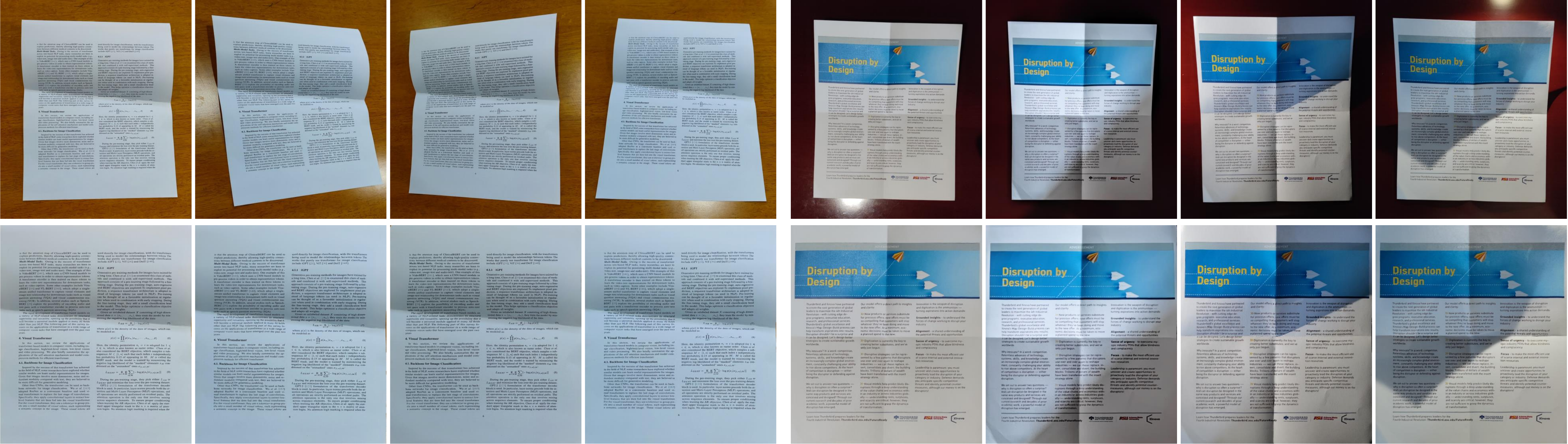}
\end{center}
\vspace{-0.18in}
\caption{Robustness illustration of DocScanner on viewpoint (left) and illumination (right) changes. 
	The two rows show input distorted and corresponding rectified images.
	The input images are captured from different viewpoints (left) and under different illumination conditions (right).}
 \vspace{-0.05in}
\label{fig:angle-ill}
\end{figure*}

\begin{figure*}[t]
\begin{center}
	\includegraphics[width=1\linewidth]{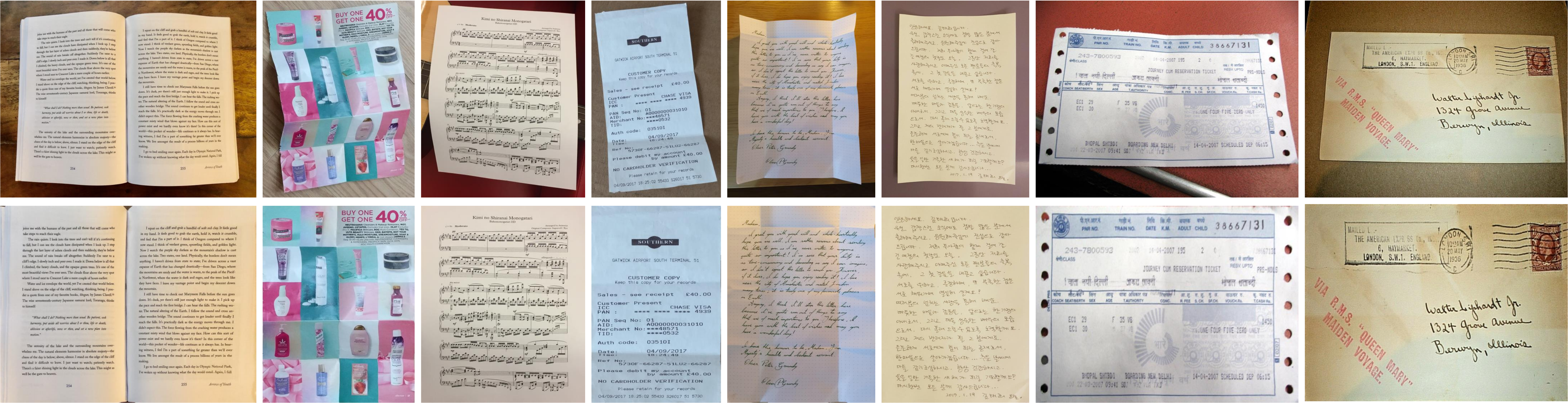}
\end{center}
\vspace{-0.18in}
\caption{Robustness illustration of DocScanner on document type changes. 
	The two rows show input distorted and corresponding rectified images.
	The types of captured deformed document include a full view of books, advertisement, music sheet, receipt, hand-written letter, ticket, and envelop, from left to right.}
\vspace{-0.1in}
\label{fig:doctype}
\end{figure*}

\smallskip
\noindent
\textbf{Comparison with state-of-the-art methods.}
As shown in Table~\ref{t1}, DocScanner sets several state-of-the-art records on DocUNet Benchmark dataset~\cite{8578592}. Here we build three varieties of our DocScanner with different model sizes (\emph{i.e.}, DocScanner-T, DocScanner-B, and DocScanner-L). Note that different from other methods, DocProj~\cite{li2019document} is a patch-based method that predicts the distortion flow on document patches rather than the entire image. Therefore, the rectified boundaries are still distorted due to the uncropped distorted images in the DocUNet Benchmark dataset~\cite{8578592}, leading to a limited performance on distortion metrics. Compared with the classic DewarpNet~\cite{9010747}, DocScanner-B achieves a relative improvement on MS-SSIM by $8.43\%$, Li-D by $18.61\%$, and CER by $24.61\%/21.41\%$, respectively, with only $1/16$ parameters. Moreover, compared with DocTr~\cite{feng2021doctr} based on the powerful transformer~\cite{Vaswani2017AttentionIA}, our larger DocScanner-L shows a relative improvement on Li-D by $11.85\%$ and CER by $10.04\%$/$14.89\%$, with $1/3$ parameters.
Compared with the recent state-of-the-art method Marior~\cite{zhang2022marior} and PaperEdge~\cite{ma2022learning}, DocScanner-L yields a sizeable improvement on metric MS-SSIM and LD.
Such lower distortion and superior OCR performance demonstrate that DocScanner can effectively restore both the structure and content of distorted document images.

As shown in Fig.~\ref{fig:freq}, we compare the distortion frequency distribution curves of DocScanner with the state-of-the-art methods~\cite{ma2022learning,jiang2022revisiting,feng2022docgeonet,zhang2022marior}. Specifically, for the Local Distortion distribution curve (left), the x-coordinate denotes the $L_2$ distance of the matched pixels between the rectified image and the GT image, while the y-coordinate denotes their frequency distribution among the total DocUNet Benchmark dataset~\cite{8578592}. We can see that, for DocScanner the pixels with small deformations take up the majority of the rectified images and the pixels with large deformations have a smaller proportion, compared to other methods.
In another word, the rectified images of DocScanner have smaller local deformations. In addition, for the Line Distortion distribution curve (right), the x-coordinate denotes the standard deviation of the rectified row and column pixels. Similarly, the y-coordinate denotes their frequency distribution among the total DocUNet Benchmark dataset~\cite{8578592}. The obtained curve (right) presents similar distributions to the Local Distortion distribution curve (left), which demonstrates that the rectified images of DocScanner have smaller global deformations.
Such results show the superior rectification performance of DocScanner over the state-of-the-art methods.

In Fig.~\ref{fig:freq_docs}, we compare the distortion frequency distribution curves of DocScanner-T, DocScanner-B, and DocScanner-L. As we can see, DocScanner-L reveals smaller local and global deformations, compared with DocScanner-T and DocScanner-B.

To better demonstrate the effectiveness of our proposed DocScanner, we further conduct qualitative comparisons with existing methods~\cite{8578592, li2019document, 9010747, xie2020dewarping,das2021end,feng2021doctr,DDCPXIE,feng2022docgeonet,zhang2022marior,jiang2022revisiting,ma2022learning}. Concretely, as shown in Fig.~\ref{fig:imgcom}, we first compare the rectified images. The results reveal that the rectified images of our DocScanner show less
distortions than the other rectification methods. Second, as shown in Fig.~\ref{fig:patchcom}, we randomly crop some local patches to compare the local rectification details. We can see that the rectified textlines of DocScanner are much straighter than other rectification methods. Such outstanding visual performances agree with the above quantitative results.

\setlength{\tabcolsep}{2mm}
\begin{table}[t]
	\small
	\centering
	\caption{Running time of processing a 1080P image and parameter count of the document localization module and the progressive rectification module.}
	\vspace{-0.08in}
	\begin{tabular}{c|cc}  
		\Xhline{2\arrayrulewidth}
		Module of DocScanner-B    & Time (s) & Parameters (M)\\   
		\hline
		document localization & 0.014 & 1.13  \\ 
		progressive rectification & 0.085 & 4.10  \\ 
		\hline
		total & 0.099 & 5.23  \\    
		\Xhline{2\arrayrulewidth}	
	\end{tabular}
	\label{para}
	\vspace{-0.1in}
\end{table}

\smallskip
\noindent
\textbf{Efficiency comparison.} As shown in Table~\ref{t1}, we also conduct efficiency comparisons, on the running time of processing a 1080P resolution image and the network parameter numbers. The evaluation is performed on a single RTX 2080Ti GPU. Note that we only compare the methods with the released codes and the computed FPS does not involve the time spent on reading images.

\begin{figure}[t]
	\begin{center}
		\includegraphics[width=1\linewidth]{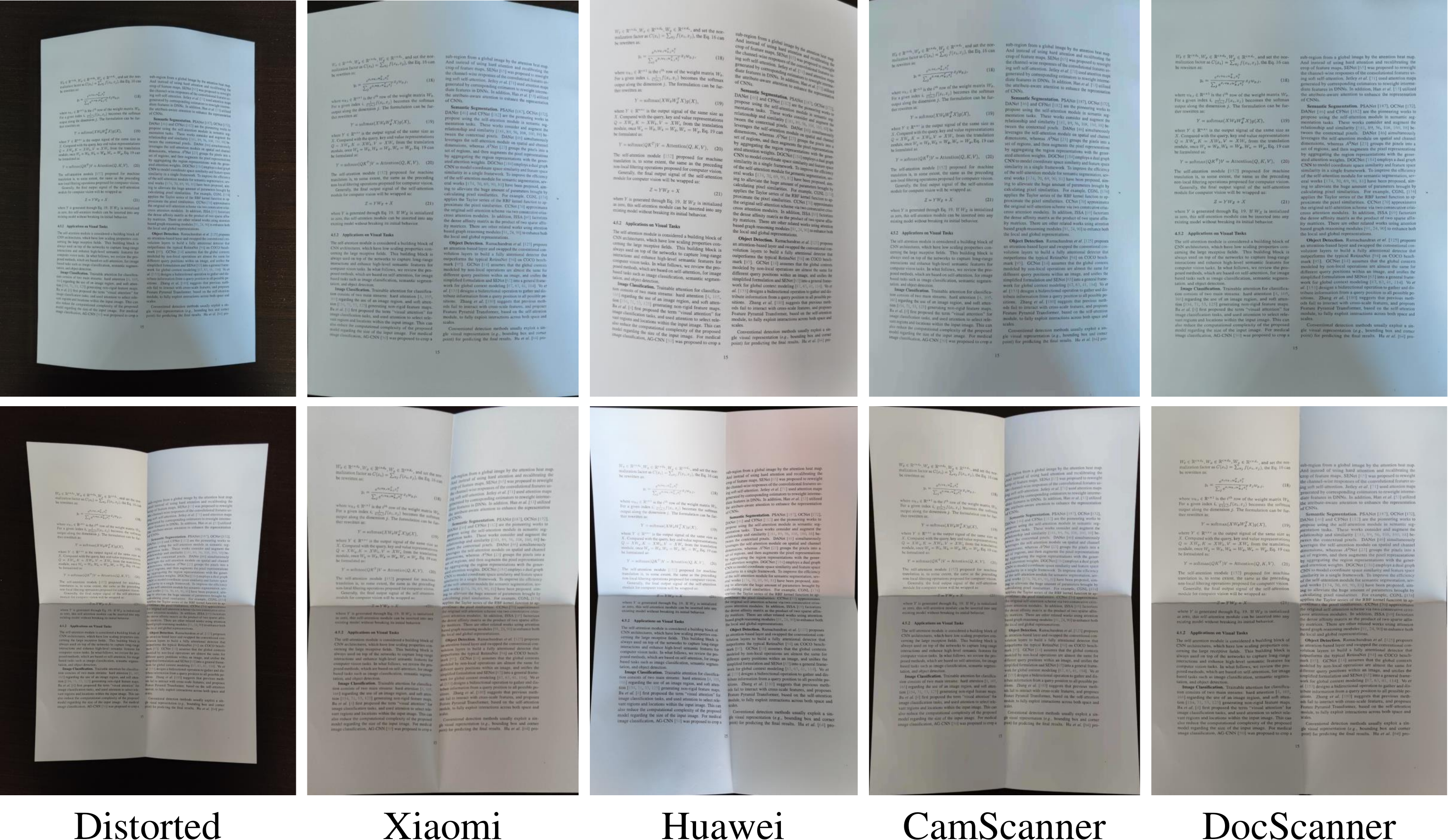}
	\end{center}
	\vspace{-0.13in}
	\caption{Qualitative comparisons with the prevalent techniques in smartphones. 
		The two rows show input distorted and corresponding rectified images.
		Different from the prevalent techniques in smartphones, our DocScanner can handle any irregular deformations.}
		\vspace{-0.1in}
	\label{fig:device}
\end{figure}

The previous method DewarpNet~\cite{9010747}, DocTr~\cite{feng2021doctr}, PaperEdge~\cite{ma2022learning}, and our DocScanner all directly predict backward warping flow for rectification. They show much higher efficiency than methods~\cite{8578592, xie2020dewarping, liu2020geometric} that take the forward warping flow as the ground truth. This is because for the latter methods, during inference, the predicted forward warping flow needs to be converted to the backward warping flow first based on scattered data interpolation~\cite{amidror2002scattered}, which is time-consuming. Besides, DocProj~\cite{li2019document} first estimates the warping flow of the image patches and then stitches them based on multilabel graph cuts~\cite{boykov2001fast}, which heavily increases the computational cost.
Compared with them, DocScanner shows superior efficiency, though it involves iterations. This could be ascribed that DocScanner applies a compact recurrent rectification module. 
DDCP~\cite{DDCPXIE} regresses a set of control points and PaperEdge~\cite{ma2022learning} estimates a sparse backward warping flow, showing higher efficiency.
Moreover, since DocScanner ties the weights across iterations, it is the most lightweight method to date. 
As shown in Table~\ref{t1}, the parameter number of DocScanner-T only has 2.6M parameters, which is approximately $3\%$ of DewarpNet~\cite{9010747} and $7\%$ of PaperEdge~\cite{ma2022learning}, and it achieves 10.81 FPS.
As shown in Table~\ref{para}, we further show the running efficiency and parameter count of the document localization module and progressive rectification module in DocScanner-B, respectively.

\setlength{\tabcolsep}{0.7mm}
\begin{table*}[t]
	\caption{Ablation experiments of DocScanner-B in terms of image similarity, distortion metrics, OCR performance, and running efficiency on the DocUNet Benchmark dataset~\cite{8578592}. ``$\uparrow$'' indicates the higher the better and ``$\downarrow$'' means the opposite.}
	\vspace{-0.08in}
	\centering
	\begin{tabular}{l|c|cc|cc|cc}  
		
		\Xhline{2\arrayrulewidth}
		\textbf{Models} &\textbf{MS-SSIM} $\uparrow$ &\textbf{LD} $\downarrow$ &\textbf{Li-D} $\downarrow$ &\textbf{ED} $\downarrow$&\textbf{CER}  $\downarrow$  &\textbf{FPS} $\uparrow$ 
		&\textbf{Para. (M)} \\  
		
		\hline
		DocScanner-B  & \textbf{0.5134} & 7.62 & 1.88 & \textbf{671.48}/\textbf{434.11} & 0.1788/0.1652  & 8.62 & 5.2 (1.1+4.1) \\
		\hline
		
		Document Localization $\rightarrow$ None & 0.4738  & 9.22 & 2.23 & 668.26/436.50 & \textbf{0.1734}/0.1668 & \textbf{10.09} & \textbf{4.1} \\
		
		Upsampling: Learnable $\rightarrow$ Bilinear & 0.5072 & 8.03 & 1.87 & 674.01/452.15 & \textbf{0.1763}/0.1679  & 8.69 & 4.8 (1.1+3.7) \\ 
		
		 ConvGRU $\rightarrow$ ConvLSTM  & 0.5131 & 7.92 & \textbf{1.86} & 684.26/448.01 & 0.1792/\textbf{0.1647} & 8.53 & 5.7 (1.1+4.6) \\ 
		
		Shared Weights $\rightarrow$ Unshared Weights  & 0.5087 & \textbf{7.52} & 1.92 & 680.62/459.52 & 0.1801/0.1693  & - & 38.7 (1.1+37.6) \\			
		
		Circle-consistency Loss $\rightarrow$ None & 0.5117 & 7.99 & 1.99 & 663.76/445.40 & 0.1787/0.1674  & - & 5.2 (1.1+4.1) \\ 
		
		\Xhline{2\arrayrulewidth}
	\end{tabular}
	\label{aba}
\end{table*}

\setlength{\tabcolsep}{1.2mm}
\begin{table*}[t]
	\vspace{-0.02in}
	\caption{Ablation experiments of the rectified feature generator of DocScanner-B in terms of image similarity, distortion metrics, OCR performance, and running efficiency on the DocUNet Benchmark dataset~\cite{8578592}. ``$\uparrow$'' indicates the higher the better and ``$\downarrow$'' means the opposite.}
	\vspace{-0.08in}
	\centering
	\begin{tabular}{c|c|c|c|c|cc|cc|cc}  
		
		\Xhline{2\arrayrulewidth}
		\multicolumn{4}{c|}{Components of ${\bm{x}}_k$}  & \multirow{2}*{\textbf{MS-SSIM} $\uparrow$} & \multirow{2}*{\textbf{LD} $\downarrow$} & \multirow{2}*{\textbf{Li-D} $\downarrow$} & \multirow{2}*{\textbf{ED} $\downarrow$} & \multirow{2}*{\textbf{CER} $\downarrow$} & \multirow{2}*{\textbf{FPS} $\uparrow$} & \multirow{2}*{\textbf{Para. (M)}} \\
		\cline{1-4}
		$\bm{c}_0$  & $Q_\theta (\bm{c}_{k-1})$ & $V_\theta (\bm{f}_m^{k-1})$ & $\bm{f}^{k-1}_m$ &  &  &  &  &  & & \\  
		
		\hline
		\scriptsize{\Checkmark} &  &  &  & 0.4736  & 9.09 & 2.70 & 1492.76/1006.88 & 0.3856/0.3687 & 9.17 & 3.9 (1.1+2.8) \\
		\scriptsize{\Checkmark} & \scriptsize{\Checkmark} &  &   & 0.4762 & 9.03 & 2.65 & 1503.82/922.05 & 0.3856/0.3645  & 9.13 & 4.7 (1.1+3.6) \\
		\scriptsize{\Checkmark} &  & \scriptsize{\Checkmark} & \scriptsize{\Checkmark}  & 0.4968 & 8.01 & 1.97 & 697.59/461.92 & \textbf{0.1741}/0.1668  & 8.70 & 4.4 (1.1+3.3) \\
		\scriptsize{\Checkmark} & \scriptsize{\Checkmark} & \scriptsize{\Checkmark} & \scriptsize{\Checkmark}  & \textbf{0.5134} & \textbf{7.62} & \textbf{1.88} & \textbf{671.48}/\textbf{434.11} & 0.1788/\textbf{0.1652}   & 8.62 & 5.2 (1.1+4.1) \\
		
				
		\Xhline{2\arrayrulewidth}
	\end{tabular}
	\label{aba:fk}
\end{table*}

\smallskip
\noindent
\textbf{Comparison with the prevalent techniques.} The prevalent algorithms built in smartphones commonly have a restriction that the document must be a regular quadrilateral. Such techniques first detect the corner points of the document to localize the document region and then perform a perspective transformation to rectify the image. Hence, these methods can not handle the situations when the captured document has any irregular deformations. As shown in Fig.~\ref{fig:device}, we compare our DocScanner with some prevalent techniques, including the CamScanner Application\footnote{https://www.camscanner.com/}, the built-in document rectification system of Huawei Mate 30 Pro, and Xiaomi 11. We can see that DocScanner is capable of correcting various irregular deformations. This is because the predicted warping flow of our DocScanner defines a non-parametric transformation, thus being able to represent a wide range of distortions.

\smallskip
\noindent
\textbf{Robustness of DocScanner.} To verify the robustness of DocScanner, we evaluate the rectification performance in four aspects, including the change of background, viewpoint, illumination, and document types.

Firstly, as shown in Fig.~\ref{fig:back}, our DocScanner can perform strongly when the captured documents are under various cluttered backgrounds.
Note that these distorted images are real document photos captured by smartphones under outdoor or indoor scenes during the day or night.
Secondly, we validate the rectification performance when the input distorted images are captured from different viewpoints and under different illumination conditions, respectively.
The results are shown in Fig.~\ref{fig:angle-ill}. 
It can be seen that DocScanner shows high robustness in spite of the various viewpoints and illumination conditions. Thirdly, as shown in Fig.~\ref{fig:doctype}, we further evaluate the ability of DocScanner to process distorted images with different document types.
The types of captured deformed documents involve a full view of book, advertisement, music sheet, receipt, hand-written letter, ticket, and envelope.
Note that such document types are blind in the training dataset but they are well-rectified by our DocScanner. These results reveal the strong generalization ability of our method.

\begin{figure}[t]
	\begin{center}
		\includegraphics[width=0.98\linewidth]{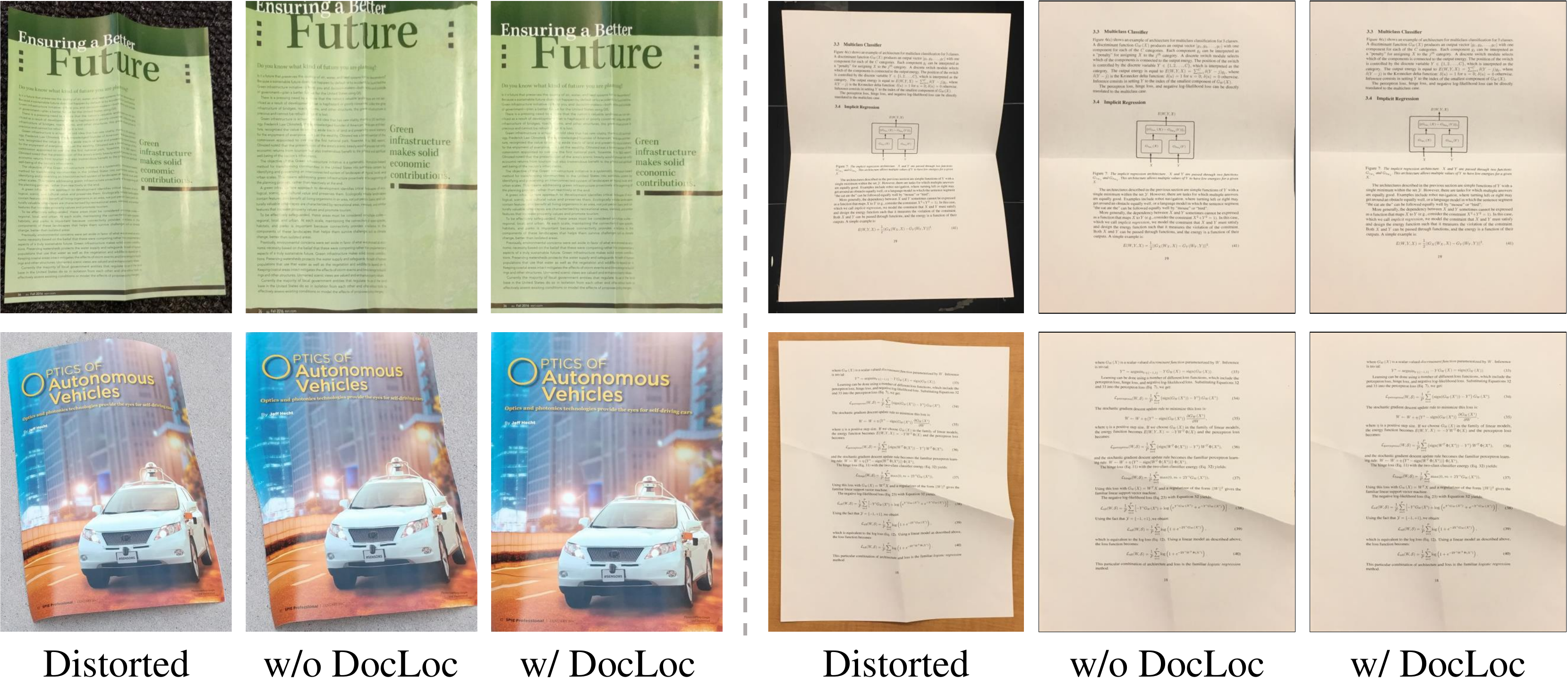}
	\end{center}
	\vspace{-0.1in}
	\caption{Examples of the results from DocScanner-B to illustrate the impact of the document localization module (abbreviated as ``DocLoc'' in this figure). The two failure cases (left) and two successful cases (right) demonstrate that document localization is auxiliary and indispensable for building a robust document image rectification system.} 
	\vspace{-0.05in}
	\label{fig:dlc}
\end{figure}

\subsection{Ablation Studies}
We conduct ablation studies to verify the effectiveness of each component in DocScanner, including the document localization module, the progressive rectification module, and the circle-consistency loss. 
Several intriguing properties are observed.

\begin{figure*}[t]
	\begin{center}
		\includegraphics[width=1\linewidth]{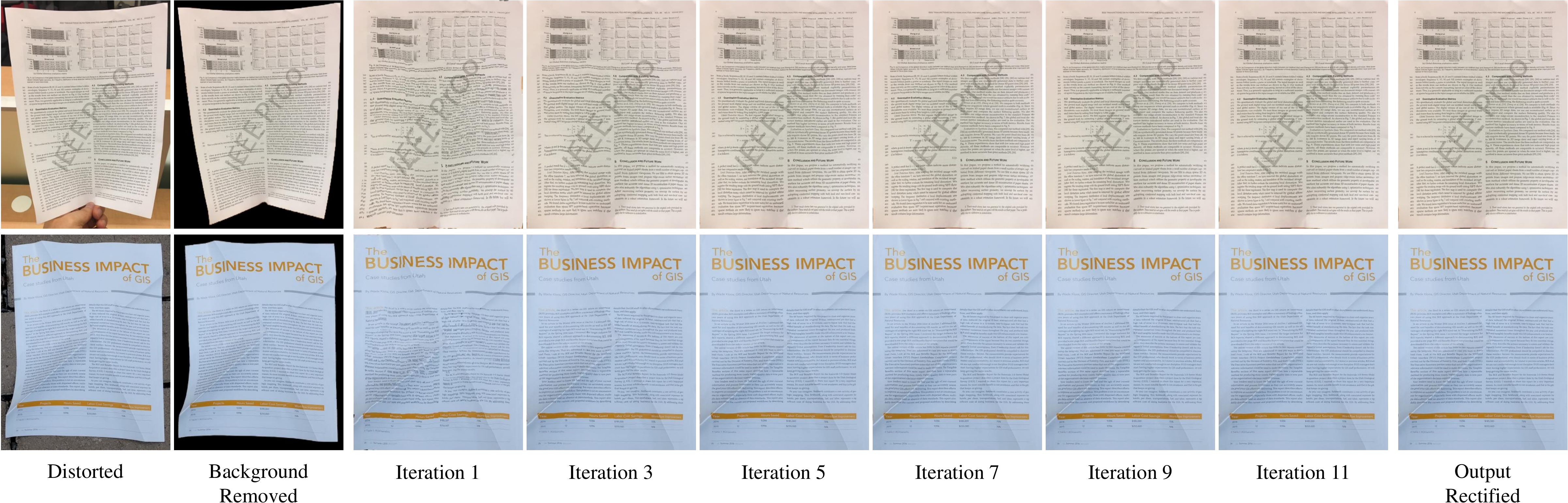}
	\end{center}
	\vspace{-0.13in}
	\caption{Visualization of the rectification process of DocScanner-B. We show the rectified document images at the selected odd iterations. It can be seen that DocScanner progressively corrects the document distortion and finally converges to a stable rectification result. It is better viewed in color.}
	\label{fig:rec_step}
\end{figure*}

\setlength{\tabcolsep}{2.5mm}
\begin{table*}[h]
	\centering
	\caption{Performance of DocScanner-B on selected iterations on the DocUNet Benchmark dataset~\cite{8578592} during inference. DocScanner does not diverge even when iteration is up to 200. Settings used in our final model are underlined. ``$\uparrow$'' indicates the higher the better and ``$\downarrow$'' means the opposite.}
	\vspace{-0.08in}
	\begin{tabular}{c|c|c|c|c|c|c|c|c|c|c|c|c|c|c}
		\Xhline{2\arrayrulewidth}		
		iters & distorted & 1 & 2 & 3 & 4 & 5 & 6 & 9 & \underline{12} & 18 & 24 & 36  & 100 & 200 \\
		\hline
		LD $\downarrow$   & 20.51      & 8.82     & 8.51      & 8.02      &  7.99     & 7.93      & 7.96    & 7.74      & 7.62   & \textbf{7.55}  & 7.57    & 7.56      & 7.60   &  7.62 \\
		\hline
		Li-D $\downarrow$   & 5.66     & 3.15     & 2.60      & 2.35      &  2.21     & 2.09      & 2.07    & 1.92      & 1.88   & \textbf{1.83}          & 1.84    & 1.86      & 1.87   &  1.87 \\
		\Xhline{2\arrayrulewidth}
	\end{tabular}
	\label{table:aba_reccurrent}
\end{table*}

\smallskip
\noindent
\textbf{Document localization module.}
Removing the noisy backgrounds or localizing the foreground document is an effective technique for improving the performance, and is widely adopted in the state-of-the-art methods~\cite{xie2020dewarping, feng2022docgeonet,zhang2022marior,jiang2022revisiting,ma2022learning} or the above built-in software in smartphones. To test its impact on our DocScanner, we train a baseline network without the document localization module, where a distorted document image with a cluttered background is directly fed to the progressive rectification module. As shown in Table~\ref{aba}, with the document localization module, the performance of DocScanner increases $17.35\%$ (from 9.22 to 7.62) and $15.70\%$ (from 2.23 to 1.88) on metric LD and Li-D, respectively. These gains can be ascribed that in DocScanner, the distorted image feature $\bm{c}_0$ is taken as the input of the warping flow updater at every iteration, as shown in Fig.~\ref{fig:2} and Fig.~\ref{fig:4}. If we do not localize the foreground documents before conducting the progressive rectification, the background noise will be injected and accumulated along with the iterations, which will disturb the rectification process.

\begin{figure}[t]
	\begin{center}
		\includegraphics[width=1\linewidth]{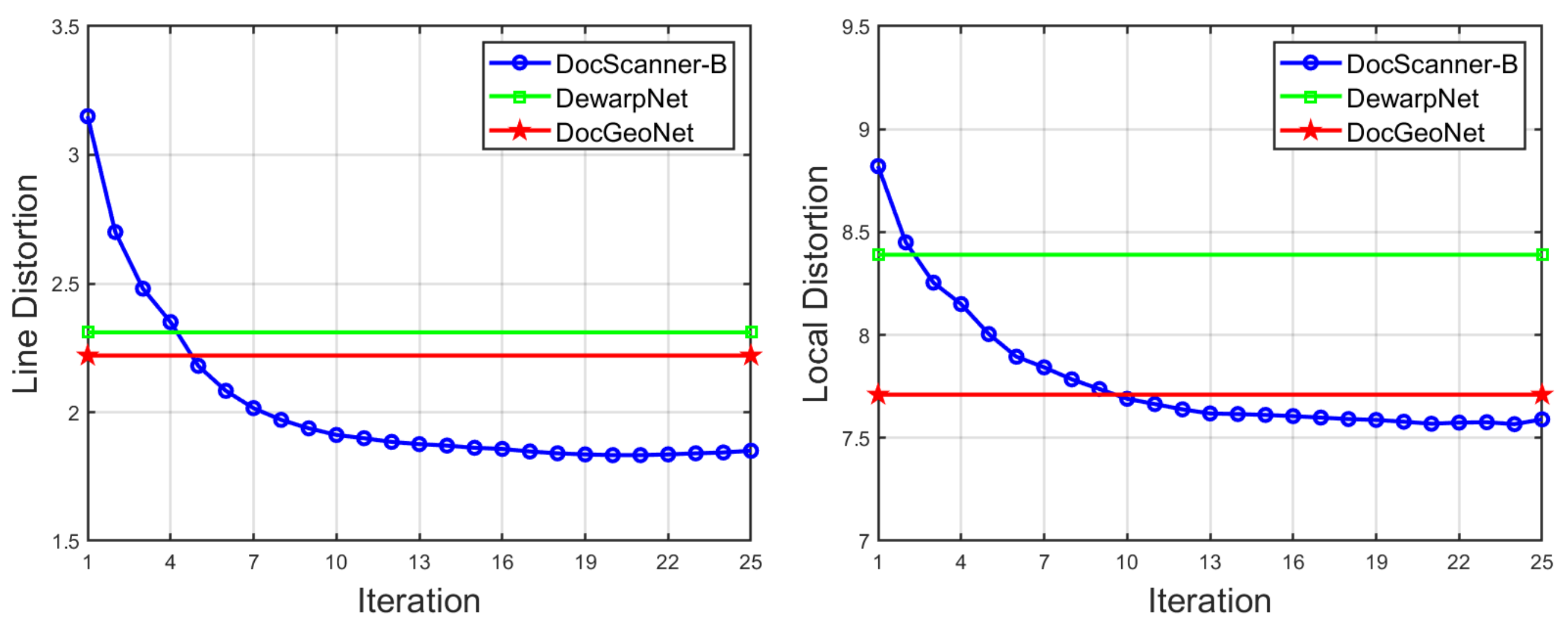}
	\end{center}
	\vspace{-0.12in}
	\caption{Performance of DocScanner-B on metric Local Distortion (left) and Line Distortion (right) from the $1^{th}$ to $25^{th}$ iteration on the DocUNet Benchmark dataset~\cite{8578592} during inference. The lower the values of LD and Li-D, the better the performance. For our DocScanner, the superior performance is obtained after convergence.}
	\vspace{-0.1in}
	\label{iter}
\end{figure}

Interestingly, as shown in Table~\ref{aba}, the improvement in OCR performance is not remarkable.
To provide a more specific view of the impact of the document localization, we showcase four examples in Fig.~\ref{fig:dlc}. As illustrated in the left two examples, without document localization, the rectified image fails to cover the complete document region (upper example), and the boundaries are corrupted with background (lower example). In contrast, the right two examples show excellent rectification quality, despite no document localization. We take a deeper analysis by counting the number of failed cases belonging to these failure situations, and find that they only account for $\bm{7.69\%}$ of the total test samples in the DocUNet Benchmark dataset~\cite{8578592}.
We observe that the images for OCR evaluation do not involve such cases, which has less influence on the OCR evaluation.
These quantitative and qualitative results verify that taking the whole distorted image as the input involves an extra burden to localize the foreground document region besides geometric rectification. More importantly, for our DocScanner, document localization is an auxiliary but indispensable part for building a robust document image rectification system.

\begin{figure*}[t]
	\begin{center}
		\includegraphics[width=1\linewidth]{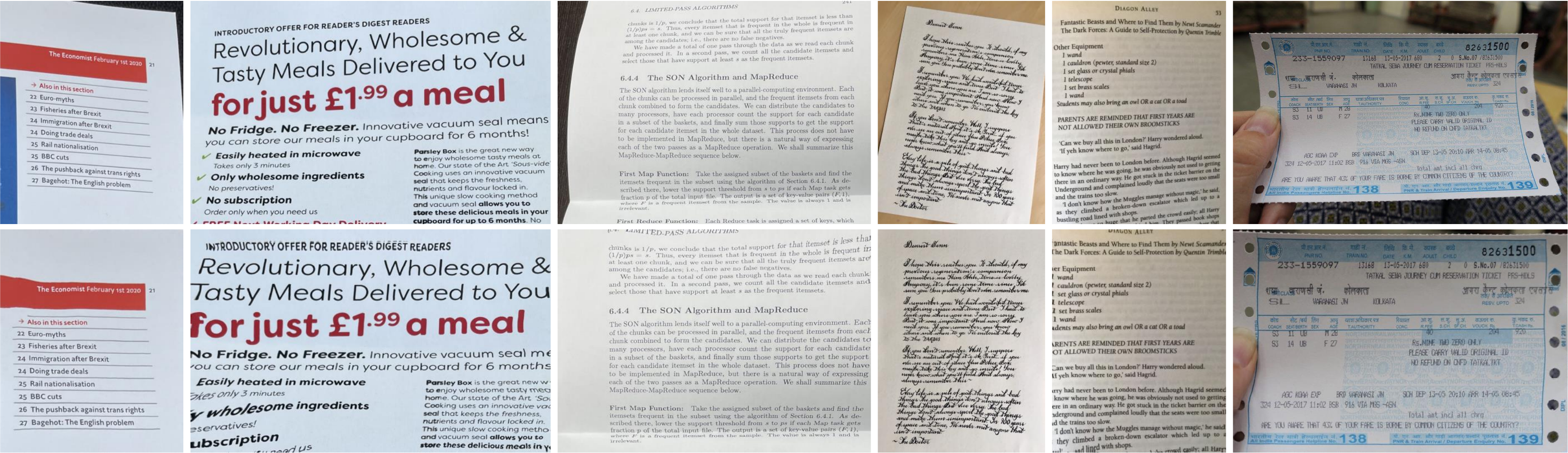}
	\end{center}
	\vspace{-0.12in}
	\caption{Example results of the limitation discussion. 
		The two rows show input distorted and corresponding rectified images of DocScanner-B, respectively.
	When the input distorted images have only incomplete or no document boundaries,
	the rectified images of DocScanner remain partial distortion.}
	\vspace{-0.12in}
	\label{limit}
\end{figure*}  

\smallskip
\noindent
\textbf{Progressive rectification module.}
In the following, we first validate the major components in the progressive rectification module, including the rectified feature generator, the learnable upsampling module for warping flow, and the warping flow updater. Then, we verify the effectiveness of our progressive learning strategy.

We first validate the compositions of the feature $\bm{x}_k$ that is fed into the warping flow updater at each iteration. Specifically, as shown in Table~\ref{aba:fk}, we train a baseline model that directly takes the distorted feature $\bm{c}_0$ as the input $\bm{x}_k$ to the warping flow updater. That is, the baseline model does not have the rectified feature generator. Then, we integrate the warped feature ($Q_\theta (\bm{c}_{k-1})$) and the flow feature ($V_\theta (\bm{f}_m^{k-1})$ and $\bm{f}^{k-1}_m$), respectively. The performances obtain a $8.8\%$ and $33.3\%$ gain on metric Line Distortion, respectively. The improvement of former ablation could be ascribed that the warped feature encodes the structure and content information of the predicted rectified image, which is differentiated and processed to estimate the further refinement. For the latter ablation, the flow feature represents the pixel displacement information, which can facilitate the learning of the residual regression for refinement. At last, DocScanner fuses the warped feature and the flow feature to generate the input feature $\bm{x}_t$. The performance gains are further enhanced due to the strong feature representations. 
 
At each iteration, the warping flow updater outputs the displacement residual $\Delta \bm{f}^k_m$ at $1/8$ resolution. Next, we compare the bilinear upsampling to our learnable upsampling module for $\Delta \bm{f}^k_m$. As shown in Table~\ref{aba}, the performances are slightly better using the learnable upsampling module. The reason is that, the coarse bilinear upsampling operation for $\Delta \bm{f}^k_m$ likely can not recover the small deformations of the distorted document.

The default updater unit in DocScanner is ConvGRU. 
We replace the ConvGRU with ConvLSTM, a modified version of standard LSTM~\cite{hochreiter1997long}. In Table~\ref{aba}, while ConvLSTM shows comparable performance, ConvGRU produces higher efficiency on inference time and parameter number.
By default, we tie the weights across the total $K$ iterations. Then, we test a version of our approach where each update operator learns a separate set of weights. Performances are slightly better when the weights are untied while the parameters significantly increase.

To provide a more specific view of the rectification process, we provide the results of the selected iteration numbers on the metric LD and Li-D in Table~\ref{table:aba_reccurrent}. The metric LD and Li-D capture the local and global distortion of the rectified document images, respectively. We can see that the main rectification lies in the top 1$ \sim $5 iterations, while the later iterations fine-tune the performance. Besides, the performance does not diverge even when the iteration number $K$ is increased to 200, which illustrates the robustness of our method. As shown in Fig.~\ref{fig:rec_step}, we further visualize the rectification process and show the corresponding rectified document images at odd iterations. It can be seen that, during the rectification process, the curved textlines in the input distorted document images are progressively corrected and finally converge to a relatively steady position, leading to a stable rectification performance. 

As shown in Fig.~\ref{iter}, we further show the performance on the DocUNet Benchmark dataset~\cite{8578592} from the $1^{th}$ to $25^{th}$ iteration during the inference process on metric LD (left) and Li-D (right), respectively. For DocScanner, the superior performance is obtained after convergence. Note that our DocScanner-B outperforms DewarpNet~\cite{9010747} after about 4 iterations, and DocGeoNet~\cite{feng2022docgeonet} after about 8 iterations. In our final model, we set the iteration number $K$=12 to stride a balance between the accuracy and the running efficiency. These quantitative and qualitative results demonstrate the effectiveness and the robustness of the progressive learning strategy.

\smallskip
\noindent
\textbf{Circle-consistency loss.}
With the circle-consistency loss, as shown in Table~\ref{aba}, DocScanner (full model) obtains an important gain on all metrics on the DocUNet Benchmark dataset~\cite{8578592}. The results illustrate the effectiveness of the straight-line constraint along the rows and columns, relieving the global distortions of the rectified document images.

\subsection{Limitation Discussion}
In this section,
we discuss the limitation of our method.
As shown in Fig.~\ref{limit},
when the input distorted images have only incomplete or no document boundaries,
the rectified images remain partial distortion.

Interestingly, 
we can see that the proposed DocScanner still shows a certain rectification capacity for such images,
though our training dataset does not contain document images with incomplete boundaries.
In fact, for a distorted document image,
its rectification cue mainly comes from three aspects,
including document boundaries, textlines, and its illumination distribution.
When document boundaries are incomplete in an image,
the rectification network still can extract the geometric information from the other aspects.

Such distorted images are also common in real life and will be explored in our future work.

\vspace{-0.02in}
\section{Conclusion}
In this work, we present DocScanner, an effective cascaded system for document image rectification. It localizes the document first and then progressively corrects the document distortion in an iterative manner.
With the progressive and iterative correction,
DocScanner achieves superior rectification performance and set several state-of-the-art scores on the challenging benchmark dataset.
Extensive experiments are conducted to validate the merits of our method.
Moreover, we propose a new distortion metric for the field that evaluate the global distortion of the rectified document images.
In the future, we will explore the rectification of document images with incomplete boundaries.
Besides, considering that DocScanner focuses on the geometric distortion problem of the document images, we intend to further concentrate on the illumination distortion to enhance the visual quality and improve the OCR accuracy. We will seek the solution in further investigations.

\bibliographystyle{spbasic}
\bibliography{reference.bib}

\end{sloppypar}
\end{document}